\documentclass[acmsmall]{acmart}

\PassOptionsToPackage{table}{xcolor} %
\usepackage{hyphenat}
\usepackage{url}
\usepackage{xspace} %
\usepackage{amsmath}
\usepackage{mathtools}
\usepackage{amsfonts}
\usepackage{algorithm}
\usepackage[noend]{algpseudocode}
\usepackage{chngcntr} %
\usepackage[show]{notes-alt}
\usepackage[table]{xcolor}
\counterwithout{algorithm}{section}
\usepackage{balance}
\usepackage{enumitem}
\usepackage{soul}
\usepackage{relsize}
\usepackage{tikz}
\usepackage{tabularx}
\usetikzlibrary{fit,positioning,calc,shapes}
\usepackage{pgfplots}
\pgfplotsset{compat=1.13}
\usepgfplotslibrary{colorbrewer}
\usepackage{multirow}
\usepackage{breqn}
\usepackage{caption}
\usepackage{pifont}
\usepackage{stfloats}
\usepackage{booktabs}
\usepackage{subcaption}

\definecolor{cycle1}{RGB}{228, 26, 28}
\definecolor{cycle2}{RGB}{55, 126, 184}
\definecolor{cycle3}{RGB}{77, 175, 74}
\definecolor{cycle4}{RGB}{152, 78, 163}
\definecolor{cycle5}{RGB}{255, 127, 0}
\definecolor{cycle6}{RGB}{153, 153, 153}%
\definecolor{cycle7}{RGB}{166, 86, 40}
\definecolor{cycle8}{RGB}{247, 129, 191}

\newcommand{\cmark}{\textcolor{cycle3}{\ding{52}}} %
\newcommand{\xmark}{\textcolor{cycle1}{\ding{56}}}
\newcommand{\mehmark}{\mbox{\cmark\textsubscript{\kern-0.45em\tiny\xmark}}}

\newcommand{\mpara}[1]{\medskip\noindent{\bf #1}}

\renewcommand{\L}{\mathcal{L}}

\DeclarePairedDelimiterX{\norm}[1]{\lVert}{\rVert}{#1}

\usepackage{xcolor}
\usepackage{framed}
\definecolor{ujwal}{rgb}{0.0, 0.5, 0.0}
\definecolor{zzj}{rgb}{0.0, 0.0, 0.7}
\definecolor{js}{rgb}{0.5, 0.0, 0.5}

\newif\ifremovecomments
\ifremovecomments
\newcommand{\ug}[1]{\relax}
\newcommand{\js}[1]{\relax}
\else
\newcommand{\ug}[1]{\textbf{\textcolor{ujwal}{UG: #1}}}%{\relax}
\newcommand{\js}[1]{\textbf{\textcolor{js}{JS: #1}}}%{\relax}
\fi

\newcommand{\dissonance}{\textsf{dissonance}}

\newcommand{\inception}{\textsc{Inception}}

\newcommand{\resnet}{\textsc{ResNet}}

\newcommand{\vgg}{\textsf{VGG}}
\newcommand{\human}{\textsf{HUMAN}}

\ifremovecomments
\newcommand{\zzj}[1]{\relax}
\else
\newcommand{\zzj}[1]
{\textbf{\textcolor{zzj}{ZZJ: #1}}}
\fi

\begin{document}
\title{Dissonance Between Human and Machine Understanding}

%\author{Authors Anonymized for Review Process}
\author{Zijian Zhang, Jaspreet Singh, Ujwal Gadiraju, Avishek Anand}
\affiliation{%
\institution{\\L3S Research Center, Leibniz Universit\"at Hannover}
 \streetaddress{}
 \city{Hannover, Germany.}
}
\email{{zzhang, singh, gadiraju, anand}@l3s.de}

% \author{Jaspreet Singh}
% \affiliation{%
%   \institution{L3S Research Center}
%   \streetaddress{Leibniz Universit\"at}
%   \city{Hannover, Germany.}
% }
% \email{singh@l3s.de}

% \author{Ujwal Gadiraju}
% \affiliation{%
%   \institution{L3S Research Center}
%   \streetaddress{Leibniz Universit\"at}
%   \city{Hannover, Germany.}
% }
% \email{gadiraju@l3s.de}

% \author{Avishek Anand}
% \affiliation{%
%   \institution{L3S Research Center}
%   \streetaddress{Leibniz Universit\"at}
%   \city{Hannover, Germany.}
% }
% \email{anand@l3s.de}

%\renewcommand{\shortauthors}{Z. Zhang, J. Singh, U. Gadiraju, A. Anand}

\begin{abstract}
%Algorithmic decision making is now prevalent in several fields including medicine, automobiles and retail. On one hand, this is testament to the ever improving performance and capabilities of complex machine learning models. On the other hand, the increased complexity has resulted in a lack of transparency and interpretability which has led to critical decision making models being deployed as functional black boxes. 

Complex machine learning models are deployed in several critical domains including healthcare and autonomous vehicles nowadays, albeit as functional blackboxes. Consequently, there has been a recent surge in interpreting decisions of such complex models in order to explain their actions to humans. Models which correspond to human interpretation of a task are more desirable in certain contexts and can help attribute liability, build trust, expose biases and in turn build better models. It is therefore crucial to understand \textit{how} and \textit{which} models conform to human understanding of tasks. In this paper we present a large-scale crowdsourcing study that reveals and quantifies the dissonance between human and machine understanding, through the lens of an image classification task.

In particular, we seek to answer the following questions: Which (well performing) complex ML models are closer to humans in their use of features to make accurate predictions? How does task difficulty affect the feature selection capability of machines in comparison to humans? Are humans consistently better at selecting features that make image recognition more accurate? Our findings have important implications on human-machine collaboration, considering that a long term goal in the field of artificial intelligence is to make machines capable of learning and reasoning like humans.

%We also explore whether humans are consistently better at selecting features that make image recognition more accurate.

%a task that humans find relatively easy and intuitive, in comparison to human recognition.

%However, there has been little work focusing if what the model understands is similar or different to that of human understanding.
%
%

\end{abstract}




\setcopyright{acmcopyright}
\acmJournal{PACMHCI}
\acmYear{2019} \acmVolume{3} \acmNumber{CSCW} \acmArticle{56} \acmMonth{11} \acmPrice{15.00}\acmDOI{10.1145/3359158}

\received{April 2019} 
\received[revised]{June 2019}
\received[accepted]{August 2019}


\maketitle

\section{Introduction}

For several decades researchers have attempted to build machine learning models that can elicit higher-order human behaviour and thinking \cite{lake2017building}. Recent advances in computational capabilities of machines alongside advances in algorithmic intelligence, have surpassed expectations and resulted in staggering feats such as `AlphaGo' defeating a world champion in the game of Go using deep neural networks \cite{silver2016mastering,silver2017mastering}. 

With all the perceived superiority of machines in decision making, arising partly from their computational prowess, we are interested in the question, ``\textit{Do machines think like humans?}.'' At the same time, it is worthy to note that humans are very good at dealing with abstract and subjective tasks, notions that machines struggle to model and cope with. This raises the question of whether humans are consistently better decision makers in tasks they are naturally suited to. 

Understanding these broad questions are crucial in building machine learning systems~\cite{Wang:2018:LCM:3219819.3220070} and guiding interpretable system design~\cite{ijcai2017-371}. More so, with the focus on algorithmic transparency where it is paramount to understand the rationale behind the decision towards building trust in the system~\cite{ieee2016ethically}. 
Intelligent machines have now become an integral part of our everyday lives, where the interaction, collaboration and cooperation between a human and an intelligent machine shapes various aspects of our society \cite{zheng2017hybrid}. Recent technological advances have led to the growing popularity of a variety of such systems, ranging from voice-based conversational assistants that facilitate and support everyday social interactions \cite{porcheron2018voice, vtyurina2018exploring}, mobile health (mHealth) applications which have been proposed to transform healthcare and for health promotion \cite{stowell2018designing}, to pervasive recommender systems which support online and offline activities of humans with growing regularity. 

There has been plenty of interest in the machine learning community towards making machines more understandable to humans, studied under \emph{interpretability of machine learning models}~\cite{kim2016examples,doshi2017towards}. One line of work focuses on building systems that are interpretable by design or whose decision process can be unambiguously explained. On the other hand there have been approaches that provide post-hoc explanations to already trained models~\cite{ribeiro2016model,lundberg2017unified}. 

To the best of our knowledge most prior work focuses largely on faithfully explaining a trained machine learning model. However little work has been done on answering the question \textit{how human-like is the machine behaving}.  A general consensus across research communities suggests that machines which can reason or act more %human-like
congruently with human expectations can create more seamless solutions for collaboration and cooperation with humans in socio-technological systems. We aim to fill this knowledge gap by enhancing the current comprehension of ``\textit{dissonance between human and machine understanding.}'' By doing so, we make important strides in CSCW and HCI towards building \textbf{machines which are more congruent with human expectations}. %more human-like machines. 
In this paper we focus on dissonance with respect to a task that is natural to humans -- image recognition~\cite{krizhevsky2012imagenet}. Our choice of task is further motivated by recent machine learning models in image classification that have reached near-human performance~\cite{szegedy2015going,szegedy2017inception}. Specifically, we focus on two scenarios of human decision making central to the image recognition task -- \emph{selection} of important parts of an image that make an object detectable in the image, and identification or \emph{recognition} of an object. The scope of this work is guided by the following research questions:

 % we should reorder these 2,1,3
 \begin{framed}
 
\begin{itemize}
\item \textbf{RQ\#1:} How do humans compare to machines in selecting important features/segments for the image classification task?

\item \textbf{RQ\#2:} What factors influence the accuracy of humans in an image recognition task?

%\item \textbf{RQ\#3:} Under what conditions can we expect a high dissonance between humans and machines?
\end{itemize}

\end{framed}

\mpara{Task in a Nutshell.} 
Towards answering these questions we employed a novel two stage crowd sourcing approach (over 7000 HITs -- human intelligence tasks) based on a consistent explanation space to gather a collective understanding of human and machine behaviour.  % add pointers to the sections

As a contextual grounding for our proposed approach to this problem, we base our task design on Biederman's theory for image understanding~\cite{biederman1985human}. The author proposed a bottom-up process, called \textit{recognition-by-components} to explain object recognition. Biederman showed that humans recognise objects by separating them into the object's main component parts. Inspired by this, we choose image super pixels as the space of input features over which we gather selection information from both humans and neural network models.
% explain biederman a bit
In the first task we ask humans to select relevant segments of an image given an object (in the image)/label that needs to be recognised. This gives us human `reasons' whereas the SHAP~\cite{lundberg2017unified} interpretability approach allows us to identify the input image segment attribution for a given decision (classified image) by a neural network. By gathering human judgements and machine explanations on the same set of segments we can directly analyse and quantify differences in reasoning which has been relatively unexplored in the literature. In the second task, we present segments of a given image one at a time to human assessors, in a decreasing order of importance determined by humans or NN models, asking them to identify the object. In doing so, we compare the dissonance between human selection versus machine selection based on the number of segments revealed towards eliciting the correct guess (i.e., the accurate class label pertaining to the given image).

\begin{figure}[h]
 	\begin{subfigure}[b]{0.3\textwidth}
    \includegraphics[width=1\textwidth]{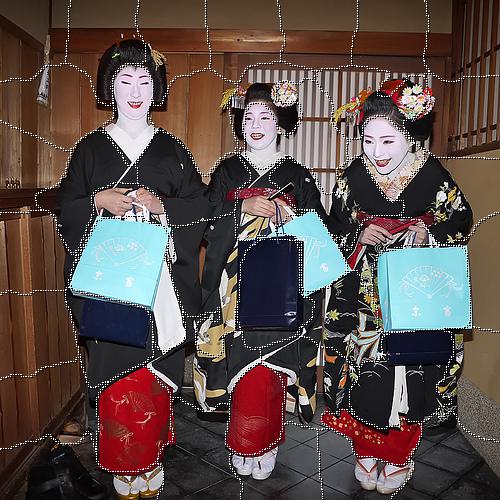}
    \caption{Original segmented image.}
	\label{fig:kimono_segmented}
    \end{subfigure}
    ~\\
 	\begin{subfigure}[b]{0.23\textwidth}
    \includegraphics[width=1\textwidth]{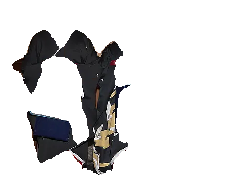}
    \caption{\human{}}
	\label{fig:kimono_human}
    \end{subfigure}
    %~
    \begin{subfigure}[b]{0.23\textwidth}
    \includegraphics[width=1\textwidth]{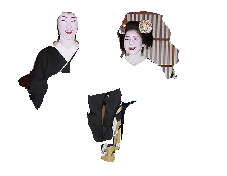}
    \caption{\inception{}}
	\label{fig:kimono_inception}
    \end{subfigure}
    %~\\
	\begin{subfigure}[b]{0.23\textwidth}
    \includegraphics[width=1\textwidth]{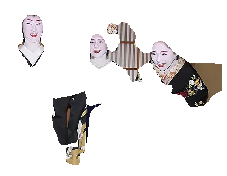}
    \caption{\resnet{}}
	\label{fig:kimono_resnet}
    \end{subfigure}
    %~
    \begin{subfigure}[b]{0.23\textwidth}
    \includegraphics[width=1\textwidth]{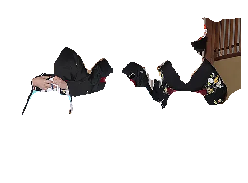}
    \caption{\vgg{}}
	\label{fig:kimono-vgg}
    \end{subfigure}
    \caption{An example of a segmented image from the `\texttt{kimono}' class (\ref{fig:kimono_segmented}) as displayed to humans in Task-1, and 5 of the most discriminative segments uncovered in Task-2 (\ref{fig:kimono_human}, \ref{fig:kimono_inception}, \ref{fig:kimono_resnet}, \ref{fig:kimono-vgg}), according to the ordering based on humans (\human{}) and machines (\inception{}, \resnet{}, \vgg{}). Humans considered the segments corresponding to the kimono itself to be most discriminative in recognizing the kimono, while the neural networks also picked contextual features such as the faces and hands of the women wearing the kimonos.}
	\label{fig:kimono-example}
\end{figure}

%Towards answering these questions we conduct a large-scale crowdsourcing study entailing 7,000 HITs (human intelligence tasks) on how similar or differently machines perceive the image classification task in comparison to humans. As a contextual grounding for our proposed approach to this problem, we base our task design on Biederman's theory for object recognition~\cite{biederman1985human} and a novel two-stage method based on segments in images. In the first task we ask humans to select relevant segments of a image given an object (in the image) that needs to be recognized. In the second task, we present segments of the image one at a time to human assessors, selected by humans or ML models, asking them to identify the object. In doing so, we compare the dissonance between human selection vs machine selection based on the number of segments shown towards the correct guess.

% As a contextual grounding for our proposed approach to this problem, we base our task design on Biederman's theory for object recognition~\cite{biederman1985human} and a novel two-stage method based on segments in images. In the first task we ask humans to select relevant segments of a image given an object (in the image) that needs to be recognized. In the second task, we present segments of the image one at a time to human assessors, selected by humans or ML models, asking them to identify the object. In doing so, we compare the dissonance between human selection vs machine selection based on the number of segments shown towards the correct guess.

\mpara{Key findings and outcomes.} A key tangible outcome is a dataset of 300 images annotated by 377 workers and 7000 HITS that we also release. Previous works have shown how human domain understanding can be utilized in building effective machine learning models~\cite{ijcai2017-371,Wang:2018:LCM:3219819.3220070}. On its own, to the best of our knowledge, this is largest dataset to be used for evaluation of interpretability for the image classification task. To ensure replicability of data collection using our tasks, the instructions for all tasks will be released along with the complete dataset\footnote{\url{https://www.l3s.de/~zzhang/cscw19}}.

% give numbers here. answer the 3 RQs in a line each. some hint of the metrics we used

%Additionally and more importantly we report some key findings of our study highlights the scenario where human understanding differs from machine understanding. 

%outperform humans in ~ 25\% more successfully recognized images
From a findings perspective, we found that neural network (NN) models that are close to human selection patterns tend to generalise well.
This has key implications on the utility of our data set towards machine learning (ML) model design. That said, our results suggest that humans do not always select the most discriminatory segments for recognition.
For example, in Figure ~\ref{fig:kimono-example}, we report the first 5 discriminative segments as perceived by humans and other ML models.
Interestingly, we find that Inception and ResNet focus on more human understandable features responsible for faster human prediction. We find that some ML models outperform humans in ~ 25\% more images. On closer examination we find that this can be attributed in part to the inability of humans to effectively choose good features from the context information that is vital for quick recognition by the crowd. Humans may potentially use more context in their decision making process than they attribute to it. Further experiments are required to fully understand this. We also use the data generated by our tasks to characterise the performance of the state-of-the-art neural networks that we chose in our study. Specifically, we find that while deeper networks tend to generalise better and choose more important features, they are less effective on difficult images. On the contrary, wide and over-parameterized networks tend to be robust in spite of being markedly different from human intuition.

Our work aims to foster research on understanding how trust manifests, builds and evolves between humans and machines, as a result of measuring the congruence of machines with human expectations. This lies at the core of HCI research, and we aim to bridge the gap between the machine learning, AI communities with the CSCW community through our work. 
%AI research is advancing at a rapid rate and has already led to society being impacted directly through facial recognition, drones, medical decision making and information prioritization. 
%It is important for the CSCW community to ensure that the societal impact of these advances are well understood before deployment. 
%This is a relatively new direction of research within CSCW, but we believe that our work can stimulate further research to address these concerns.

\section{Background and Related Work}
\label{sec:relwork}

Researchers in the CSCW and HCI communities have shed light on the unintended consequences of algorithms and machine learning models that can have a societal impact that is unanticipated by their creators \cite{Binns:2018:RHP:3173574.3173951, o2016weapons}. Others have also reflected on the benefits that machine learning models can offer to the society at large by supporting human decision-making \cite{anand2018effects, kahneman2016noise, kleinberg2017human}. Several machine learning models mediate our social, cultural, economic and political interactions in today's world \cite{rahwan2019machine}. Therefore, understanding these models and how congruent they are with human expectations is of paramount importance, so as to control their actions, enjoy their benefits and mitigate their harms. For example, online pricing models have been shown to shape the cost of products differently to different customers \cite{hannak2017bias}. Understanding the full breadth of societal effects that machines can have becomes more complex in hybrid systems composed of many humans and machines interacting; demonstrating collective behaviour \cite{shirado2017locally}. In a recently laid out HCI research agenda, authors reflected on how a lot of work in the AI and ML communities tends to suffer from a lack of usability, practical interpretability and
efficacy on real users, calling the HCI community to take the lead to ensure that new intelligent systems and ML models are transparent from the ground up, and congruent to human expectations \cite{abdul2018trends}.
In this paper, we aim to bridge the knowledge gap in understanding how congruent machine learning models are with the expectations of humans in image classification tasks, where machine learning models have been shown to be on par with human performance. Our findings have direct implications on HCI and CSCW research that aims to understand how humans and machines differ in their decision-making. We make a foundational contribution towards studying the decision-making processes of humans and machines, attempting to understand how and where they differ.

We discuss related literature in four broad realms -- (1) work on algorithmic transparency by using explanations understandable to humans, (2) methodological approaches in model interpretability, (3) neuroscience approaches that explore the `humans versus machines' context in object recognition, and (4) theories on human understanding.

%We classify the related work into three broad topics -- (1) work on algorithmic transparency by explanations understandable to humans, (2) more methodological approaches in model interpretability, and (3) theories on human understanding. 
%\mpara{Human Cognition and ML Tasks.}
 
%\mpara{Interpretability in HCI} In the HCI community~\cite{Rader:2018:EMS:3173574.3173677, Binns:2018:RHP:3173574.3173951,Veale:2018:FAD:3173574.3174014} interpretability has also been investigated ....

\subsection{Algorithmic Transparency and Explanations} Today's world is characterised by an increasing dependency on algorithmic decision-making systems \cite{zarsky2016trouble}. Since these systems augment our everyday lives, recent CSCW and HCI research has reflected upon the importance for people to understand them better \cite{abdul2018trends}.
 As described by Rader et al., algorithmic transparency involves encountering non-obvious information that is typically difficult for the user of a system to learn and experience directly, about \textit{how} and \textit{why} a system works the way it does and \textit{what} this means for the system's outputs \cite{Rader:2018:EMS:3173574.3173677}. 
 Several recommender systems provide explanations alongside their recommendations with an aim to be more persuasive, ensuring that the system's goals are served \cite{berkovsky2017recommend}. Explanations in such contexts present a user with information regarding how and why the system produced a given recommendation. Prior works have focused on various attributes of explanations; cognitive fit \cite{giboney2015user}, content type \cite{gregor1999explanations}, data sources \cite{papadimitriou2012generalized}, and modality \cite{oduor2008effects}. In other work, authors classified explanations into `black box' and `white box' descriptions \cite{friedrich2011taxonomy}. `Black box' explanations provide justifications for the outcomes of a system but do not disclose and discuss how the system works \cite{wang2007recommendation}. On the other hand, `white box' explanations delve into the inputs and outputs of a system and the steps taken through the course of arriving at particular outcomes \cite{tintarev2007survey}. Recent work by Binns et al. argued that there may be no `best' approach to explaining algorithmic decisions \cite{Binns:2018:RHP:3173574.3173951}.

%CSCW work on algorithmic fairness in the sharing economy \cite{lee2015working}, and algorithmic mediation in group decisions \cite{lee2017algorithmic} have reflected on the importance of algorithmic transparency. 

A significant amount of prior work has focused on the importance and effects of algorithmic transparency and the role of explanations to help human users comprehend the functioning of intelligent machines better. This includes work from the CSCW community on algorithmic fairness in the sharing economy \cite{lee2015working}, and algorithmic mediation in group decisions \cite{lee2017algorithmic}. However, few works have juxtaposed human understanding with that of machines. In this paper, we aim to fill this gap by studying the dissonance between human and machine understanding.

\subsection{Interpretability in Machine Learning} 

% Interpretability in Machine Learning has been studied for a long time in classical machine learning. However, the success of Neural networks (NN) and other expressive yet complex ML models have only intensified this discussion further.

% Most of the success of NN models arise from the fact that they are powerful function approximators and have succeeded in learning complex non-linear hidden representations from raw input high-dimensional representations. 

% On one hand this has largely improved the performance of ML models, but on the other they tend to be opaque and less interpretable. Consequently, interpretability of these complex models has been studied in various other domains to better understand decisions made by the network -- image classification and captioning~\cite{captioningxu2015showattention,dabkowski2017real,imageclasssimonyan2013deep}, sequence to sequence modeling~\cite{seqalvarez2017causal,nlp2015visualizing}, recommender systems~\cite{interpretrecsysChang:2016} and so forth. 
Unlike work on creating explanations it's important to note that there's a difference between explaining why a system behaves a certain way and \emph{interpreting a model}.
Interpretable models can be categorised into two broad classes: \emph{model introspective} and \emph{model agnostic}. Model introspection refers to ``interpretable'' models, such as decision trees, rules~\cite{rulesletham2015interpretable}, additive models~\cite{caruana2015intelligibletrees} and attention-based networks~\cite{captioningxu2015showattention}. Instead of supporting models that are functionally black-boxes, such as an arbitrary neural network or random forests with thousands of trees, these approaches use models in which there is the possibility of meaningfully inspecting model components directly, e.g. a path in a decision tree, a single rule, or the weight of a specific feature in a linear model. 

Model agnostic approaches on the other hand extract post-hoc explanations by treating the original model as a black box either by learning from the output of the black box model, or perturbing the inputs, or both~\cite{ribeiro2016should,influencefunctionskoh2017understanding}. Model agnostic interpretability is of two types: local and global. \emph{Local interpretability} refers to the explanations used to describe a single decision of the model. There are also other notions of interpretability, and for a more comprehensive description of the approaches we point the readers to~\cite{lipton2016mythos}.

Local Interpretability can be model agnostic or introspective. In the model agnostic case like in~
\cite{ribeiro2016should}, a simple linear model is trained to explain a single data by perturbing the data point systematically and labelling the new synthetic data using the model. 

%On of the most popular model agnostic approach is  LIME contributed by Marco et al. in \cite{ribeiro2016should}. Instead of inspecting mechanism of models making prediction, the LIME cares about the \textit{local fidelity} of the model only. 

%\ug{Explain what this means in a sentence or two - CSCW readership.} 

% Roughly speaking, one test data point together with some of its additional perturbation form a neighborhood. The section of the dataset's decision boundary intersected by this neighborhood is approximately linear, i.e. the boundary is local linear. The LIME tries to construct an interpretable linear model, which classify the neighborhood in the same way as the original model does.
More recently, Lunderberg and Lee~\cite{lundberg2017unified} introduced their model introspective approach, also known as SHAP, which utilizes the classical Shapley value estimation method from cooperative game theory. In essence, SHAP generates feature importance values for a given decision over a pre-trained model by propagating differences in activation to the expected value through the network. In this work, we use SHAP scores over the image segments (that we consider as features in our setting) to compute feature importance in Task-2.

% All explanation approaches like LIME mentioned above and the LRP to be introduced in the following, are categorized by S.Lunderberg and S.Lee~\cite{lundberg2017unified} into the \emph{additive feature attribution methods}. They also generalized such methods by establishing three desirable properties -- local accurac. These three properties are:

% \begin{itemize}
% \item \emph{Local accuracy}. The simpler, mostly linear explanation model produces the same prediction as the original, complex, mostly non-linear model does, when they are fed with a simplified input as well as the corresponding original input.
% \item \emph{Missingess}. In such methods presence of some original features are represented by corresponding simplified features. The missing features in the original input involve the absence of impact of their corresponding simplified feature.
% \item \emph{Consistency}. The attribution of a simplified feature is positive related with the contribution of its corresponding original features.
% \end{itemize}
% Other additive methods vitiate \emph{consistency} or \emph{local accuracy}, while the only one additive feature attribution method respecting all three method is the SHAP introduced by S.Lunderberg and S.Lee.

\subsection{Humans versus Machines : Neuroscience Approaches}

Our work in this paper is not the first attempt to study how humans and artificial neural network (NN) models differ in the way they perceive objects. Afraz et al. proposed falsifiable, predictive models that account for neural encoding and decoding processes that underlie visual object recognition \cite{afraz2014neural}. With an aim to better understand neural encoding in the higher areas of the ventral stream\footnote{The ventral stream is involved with object and visual identification and recognition (cf. the two-stream hypothesis \cite{eysenck2013cognitive}).} of human brains, Yamins et al. used computational techniques to identify a NN model that matches human performance on an object categorisation task \cite{yamins2014performance}. Authors found that the model was highly predictive of neural responses in both the V4 cortex and the inferior temporal cortex, the top two layers of ventral visual hierarchy in humans. Schrimpf et al. proposed \textit{Brain-Score}, a composite of several neural and behavioural benchmarks that score a neural network on how similar it is to a primate brain's mechanisms for core object recognition\footnote{\textit{Core object recognition} is the ability to rapidly recognise objects despite variations in their appearance.} \cite{schrimpf2018brain}. Rajalingham et al. systematically compared specific neural network models with the behavioral responses of humans and monkeys at the resolution of individual images \cite{rajalingham2018large}. The authors found that the NN models which they tested, significantly diverged from primate behavior.

In contrast to the aforementioned approaches that utilize fMRI's and other sensing devices to correlate features with NN models, in this work we rely on gathering explicit feedback from humans on their decision-making process for the task of object recognition. Although object recognition is intuitive to humans, understanding reasons for their decisions in unobtrusive ways (for example, by using eye tracking, fMRIs, etc.) is expensive and does not scale easily. The novelty of our work lies in understanding dissonance between humans and machines based on instance-level fine grained reasoning due to our choice of task, NNs and interpretability techniques.
%By gathering and aggregating multiple human judgements on segment importance we believe that the resulting order sheds enough light (for this work) on why a human would recognize a given object.

%\cite{rajalingham2018large,kar2018evidence,schrimpf2018brain,yamins2014performance,afraz2014neural}

%Granted these papers take a neuroscience approach to studying these problems, this paper should not only cite these papers but should also trade off their approach, which asks for subjective responses from crowd workers versus the research done in these papers that is arguable more objective as they utilize fMRI’s and other sensing devices to correlate features with neural networks.

\subsection{Human Understanding and Intuition}

Cognitive scientists have proposed that much of our thinking, memory and attitudes all operate on two levels: conscious and deliberate, and unconscious and automatic~\cite{myers2002powers}. Intuition is our capacity for immediate insight without observation or reason, i.e. thinking without conscious awareness. Kahneman~\cite{kahneman2003perspective} argues that like the perceptual system, intuition operates through impressions and judgements that directly reflect impressions.  %can be called intuitive.
In contrast, deliberate thinking is reflective, reasoning-like, critical, analytic and operates in the realm of conscious awareness. 
Intuitive judgements can of course be overridden by a more deliberate, rational process but intuition may still affect subsequent responses through priming~\cite{kahneman2003perspective}.

% Intuitive judgements are more or less accessible to an individual depending on a number of factors including: physical properties of the object of judgement (e.g. those which are routinely registered by the perceptual system such as size, distance, loudness and similarity); physical salience (e.g. larger items are more visually salient); priming through associative activation (i.e. contextual framing) as well as emotional and motivational states~\cite{kahneman2003perspective}. Intuitive judgements can be overridden by a more deliberate, rational mode of operation. However, they may still affect subsequent responses through priming.
Consequently, human decision making is based on these two levels of rationality. While even the most tedious decisions that appear to be deliberate and well considered like market investments or medical diagnostics involve a certain amount of intuition. Herbert Simon's theory of bounded rationality~\cite{simon1997models} argues against the strict rationality model and states that decisions can be made with reasonable amounts of calculation, and using \emph{incomplete information}. 

With an aim to further the understanding of human-machine dissonance, we chose the machine learning task of \emph{image classification}, since humans are known to be capable of solving image recognition tasks with high accuracy using their intuition and deliberation. 
%We chose simple object recognition in images since (1) humans are adept at it, requiring little effort & deliberation,
Moreover, neural networks (NNs) have matched and surpassed human performance on many benchmarks in the task of object recognition and are being used in various real-world applications~\cite{szegedy2017inception,szegedy2016rethinking}. This task also has added benefits from a feasibility standpoint -- several trained NNs with clear descriptions of their architecture are freely available. Interpretability techniques developed in the machine learning community allow us to examine the decision making process of NNs. Having been studied over several years for object recognition in particular, these interpretability techniques are now mature. %Through the lens of the image classification task
Towards this end, we involve a large number of human subjects %with an inherently varying degree of expertise
in a crowdsourcing setting, as described in the following section.

%More so since 

%\section{Which neural network is closest to Human Intuition?}

\section{Study Design}
\label{sec:study_design}
%\subsection{Human Image Understanding}
%\ug{Discuss Biederman's work} \cite{biederman1985human}.

\subsection{Data set Description}

The ImageNet data set was created to help train machine learning models classify objects in images \cite{deng2009imagenet}. It consists of over a million images and 1000 classes. Each image is labelled with a single class even if there are multiple objects in the image. Classes range from broad categories like `\texttt{minivan}' to specific breeds of dogs like `\texttt{shih-tzu}'. As motivated by prior work, creating ground truth data for evaluation using human input and intuition is often an expensive process when scaled \cite{crowdscale2013}. This is indeed the case for industry-sized data sets such as ImageNet \cite{krishna2016embracing}. Moreover, to study the research questions posed earlier we are not constrained by a need for a very large data set. Thus, we selected 50 classes out of 1000 and sampled 6 images at random from each class to create a data set of 300 images. Additionally we also ensure that all chosen images are classified correctly by the models we consider. 

%Creating a ground truth for human intuition is expensive for the whole of ImageNet and unnecessary to study the research questions posed earlier. Instead we selected 50 classes out of 1000 and sampled 6 images at random from each class to create a data set of 300 images. 

We solicited the aid of 3 researchers in our university to select these 50 classes pro bono. We only showed them the full list of classes (not the images).
% , and listed the classes in the decreasing order \zzj{might need to clarify the 'decreasing order'}of corresponding images contained in ImageNet. 
We defined selection criteria based on the scope of our research as follows: 
\begin{itemize}[leftmargin=*]
\item \textbf{Familiar}: the class should be familiar to all the annotators, i.e., all annotators should know what exactly the selected class of objects refers to. This criteria was added to help select classes that most people would recognise and reduce undue effort from crowd workers. 

\item \textbf{Unambiguous}: the class should have only one clear connotation for the given object. For instance, the class `\texttt{crane}' can refer to either the machine or the animal, and is thereby ambiguous.
\item \textbf{Non-specific}: the class should not be a specialisation or a potential sub-class of another class in ImageNet. If it is then neither class can be selected. For example, the classes `\texttt{cat}' and `\texttt{Persian cat}'. Since crowd workers are not experts in identifying various fine-grained classes of objects, we cannot expect them to be able to identify features pertaining to a very specific class whereas the ML models are exposed to all classes in training.
 
\end{itemize}

Apart from this, we also gathered annotations based on whether the annotators believed that it would be easy to identify objects from the selected class in a given image. We marked classes as \textit{difficult} to identify if at least one annotator indicated so. From this process we ended up with 28 \textit{easy} classes and 22 \textit{difficult} classes. 
Finally, we considered the first 50 classes that the annotators completely agreed on according to the criteria.

\subsection{Neural Networks for Object Classification}
We employed three neural networks in our experiments -- VGG19 \cite{simonyan2014very}, Inception-ResNet-V2 \cite{szegedy2017inception} and Inception-V3 \cite{szegedy2016rethinking}. These are state-of-the-art models that report high accuracy and human-level performance on the ImageNet data set. Furthermore, they differ in key areas of their network architecture which is discussed below.

%\ug{Rephrase the last sentence here. We need to describe the rationale behind selecting these NNs better.} \zzj{They are all state-of-the-art models performing the same as or superior than human does on image classification tasks. And all three of them are deep neural model and thence are used as black-boxes.}

First released in 2014, VGG19 won the first and second prize of ILSVRC (ImageNet) localisation and classification challenges. It has 16 convolution layers and 3 fully connected layers that made it one of the deepest NN architectures at the time. They report a 74.5\% Top-1 accuracy on the validation data of ILSVRC2012 \cite{deng2009imagenet} contest. The number of parameters (143,667,240) of VGG19 is the highest among the three models chosen in this work.

Inception-V3 is an improved version of the original GoogLeNet \cite{szegedy2015going}. They introduced concatenated pooling layers and showed that breaking down the large convolution kernels into several small ones significantly improved the performance as well as reducing the number of parameters. The number of parameters (23,851,784) is the smallest among the three models chosen, and the Top-1 accuracy reported is 78.8\%.

%As the depth of neural networks increases, the training stability of the nets gets worse. When training using the Back Propagation strategy, the gradient may even vanish. By adding residual units, the Inception-V4, Inception-ResNet-V1 and

Inception-ResNet-V2 has a hybrid structure consisting of residual and inception units that accelerate the training while maintaining the precision of the network. The depth of Inception-ResNet-V2 is 572 and the highest among three models considered in this work, while its number of parameters (55,873,736) is approximately two times that of the Inception-V3. Its Top-1 accuracy on ILSVRC2012 is 82.2\%.
% \todo{Zijian can fill this in. 3 lines in general about NN's and then short descriptions for each network. we could even replace this with a table. important to point out that all networks perform nearly the same on the test set.}

For the sake of readability we will refer to the VGG19, Inception-ResNet-V2, and the Inception-V3 models as \vgg{}, \resnet{}, and \inception{} respectively hereafter in this paper. 

\subsection{Method}
\label{sec:method}

Models that make decisions close to the way humans do are often desired and tend to perform less perplexing-ly on unseen data \cite{ijcai2017-371}. Even if models have similar performance according to metrics like \textit{accuracy}, they may differ in terms of the reasons that drive the making of their decisions. These reasons can be attributed to the training data, architecture, training procedure or a combination of such factors. In this work we focus on models that have been trained and validated using the same data but have different architectures; all three neural networks (\vgg{}, \resnet{}, and \inception{}) were trained on the same 1.2M images belonging to 1K classes in the ImageNet data set.

Our task design is inspired by Biederman's seminal work on human image understanding \cite{biederman1985human}. Biederman proposed the \textit{recognition-by-components} theory, which can account for the major phenomena of object recognition. He showed that if an arrangement of a few primitive components can be recovered from the input, the objects can be quickly recognised even in the presence of a significant amount of noise. Thus, in the context of object recognition in images, we define human intuition or reasoning in terms of the segments in an image which are perceived to aid the accurate recognition of the image class or label.

\begin{figure*}
 	\begin{subfigure}[b]{0.33\textwidth}
    \includegraphics[width=1\textwidth]{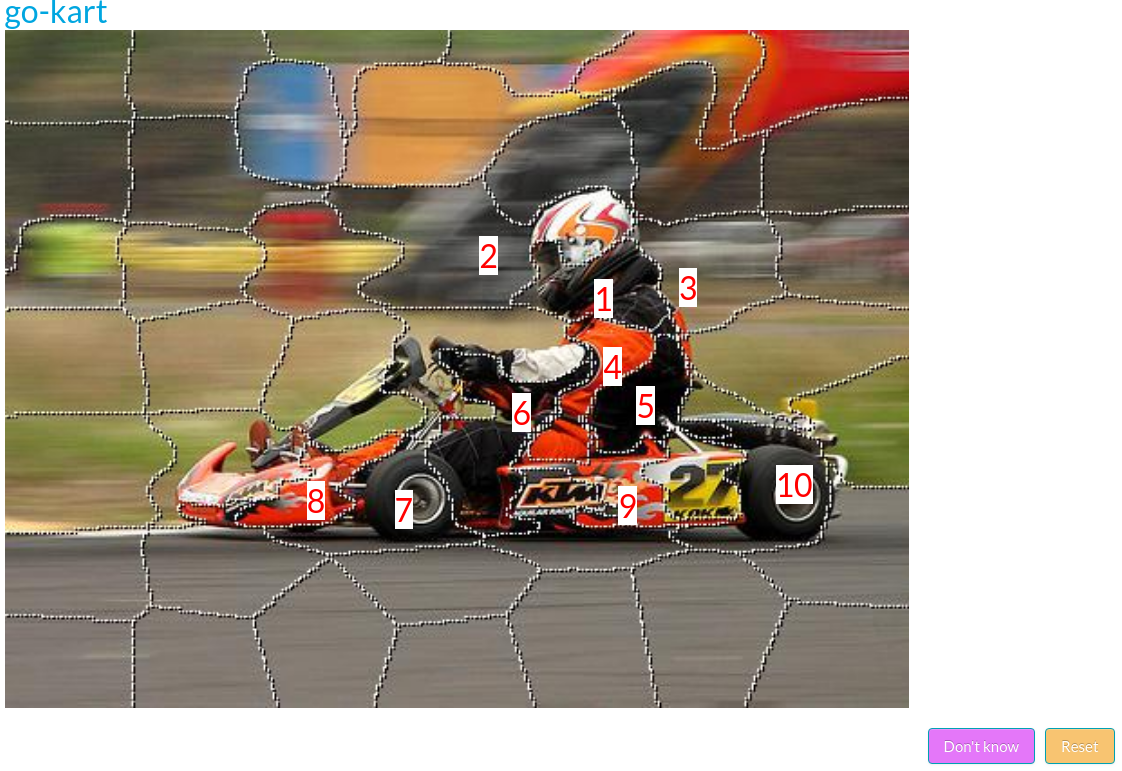}
    \caption{Task 1}
	\label{fig:task_1_example}
    \end{subfigure}
    ~
    \begin{subfigure}[b]{0.33\textwidth}
    \includegraphics[width=1\textwidth]{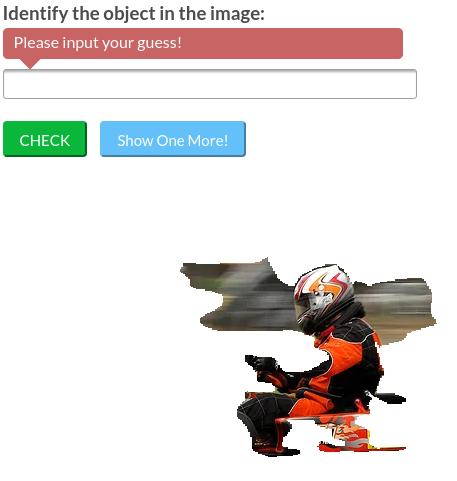}
    \caption{Task 2 - Human Segments}
	\label{fig:segs_all_diff}
    \end{subfigure}
    ~
	\label{fig:segs_all_results}
	\begin{subfigure}[b]{0.33\textwidth}
    \includegraphics[width=1\textwidth]{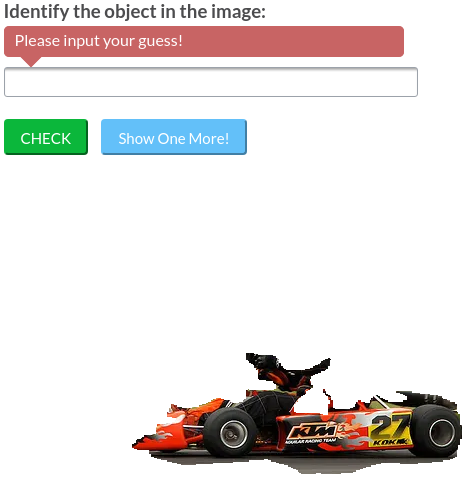}
    \caption{Task 2 - Machine Segments}
	\label{fig:segs_all_stacked}
    \end{subfigure}
    ~
    \caption{Tasks in our Crowdsourcing study. (a) Task 1 presents the clickable-segmented image along with the actual object name. (b) and (c) show the image recognition UI for Task 2 where segments are shown one at a time. The initial machine selected segments is shown in (b) and the human selected images is shown in (c) for the same image class \texttt{go-kart}.}
	\label{fig:task-design}
\end{figure*}

For instance, take the Figure shown in \ref{fig:task_1_example} whose class label according to ImageNet is ``\texttt{go-kart}''. To correctly identify the object as a \texttt{go-kart}, not only are the segments corresponding to the kart strong reasons but so are those pertaining to the driver. To accurately capture human intuition in this task, we not only need all the segments that humans use to make a decision but we also need to understand the relative importance of each segment. 
To this end we first deployed a crowdsourcing image classification task on FigureEight\footnote{http://www.figure-eight.com/}, a primary crowdsourcing platform, to gather human intuition judgements corresponding to the 300 images from 50 different classes. 

\subsubsection{Crowdsourcing Task Design -- Image Classification (Task-1)}
\label{sec:task-1}

We divide each image into 50 segments. %The segmentation is done by clustering pixels in the image based on their RGB values. 
Super pixel segmentation is utilised to cluster spatially similar pixels into a fixed number of segments. Standard grid lines also allow for such fixed size segmentation but are unaware of object boundaries, which is crucial in identifying segments of importance. For example, a single segment in a grid can contain an important part of the object and a large part of the background which may be non-essential. Super pixels are less susceptible to such effects and are hence also utilised by SHAP and other approaches like LIME \cite{ribeiro2016should}. 

Crowd workers are shown the images and their corresponding labels, and then instructed to select all segments in the image that help them correctly identify the given object. %(irrespective of its position in the image)
Workers are urged to select segments in the order of perceived importance, where the first segment they select is the strongest indicator of the object in the image. Annotators can click on each segment to select it. The first segment that is clicked is marked with the number 1 and every subsequent click is also recorded and displayed with the corresponding selection number as shown in the Figure \ref{fig:task_1_example}. %Annotators were asked to select a minimum of 3 segments per image. 
Note that the workers were explicitly encouraged to select the most important segments that could help in identifying the object in the image, including segments with contextual cues.

% task meta details like cost, test questions and time taken on average goes here
We collected 5 distinct judgements for each of the 300 images. Workers were paid at an hourly rate of 7,50 USD. To ensure a high reliability of judgements gathered, we restricted participation to the highest quality workers using an inbuilt feature on the platform\footnote{\textit{Level-3 contributors} on FigureEight comprise workers who completed $>100$ test questions across hundreds of different types of tasks, and have a near perfect overall accuracy.}. We created gold-standard data and used test questions within the task, facilitating training of workers and maintaining the overall quality simultaneously \cite{gadiraju2015training,oleson2011programmatic}. We balanced the distribution of \textit{easy} and \textit{difficult} classes in our gold-standard data by having an equal number of images from the \textit{easy} and \textit{difficult} classes to prevent potential biases. For convenience, we will refer to this task as Task-1 hereafter. Through the remainder of the paper, we do not use the terms `easy' and `difficult' to refer to the classes. We define image level difficulty as perceived by workers in Section \ref{sec:diff-diss}.

\subsubsection{Rank Agreement and Aggregation}
\label{sec:agreement-aggregation}

In our setting, humans select image segments in order to help identify an object. The human annotations are essentially an ordering or ranking of image segments per image per crowd worker. Rankings are inherently different from categorical and ordinal scale annotations which means we cannot use standard agreement measures (like Fleiss' Kappa) or aggregation methods like average or majority voting. 

In our case, since we do not enforce an exact number of segments to select, we have non-conjoint partial rankings, i.e. for the same image we can have (i) different segments (ii) a varied number of segments (iii) differing preferences. Additionally we're most interested in the top ranked segments. Standard rank correlation metrics like Kendall's Tau are not designed to handle these conditions. 
A better measure for this purpose is \emph{rank biased overlap} (RBO)~\cite{Webber:2010:RBO} that is specifically designed to address these shortcomings. 
To measure the segment selection agreement between workers for an image we compute the average pairwise RBO. In our experiments we found a high agreement between workers, with $RBO=0.7$. 

Once we have the rankings over segments from each human annotator we need to aggregate multiple rankings for the same image into one aggregated ranking.
Aggregating rankings is a well studied problem. 
The Placket-Luce (PL)~\cite{plackett1975analysis} model is particularly used for partial rankings. The PL model is a k-way rank aggregation model that is a generalised case of a parametric choice model, known as the Bradley-Terry model meant for the case of pairwise comparisons. 
Given a set of rankings we estimate the parameters of the model using maximum likelihood. Each parameter corresponds to the probability of selection for an item from a set of alternatives. 
We order the segments  based on the \emph{estimated PL model for each image}. For segments that are not selected by any workers we randomise their order and append them to the list of ordered segments. We then convert the parameter estimates for the segments into a probability distribution using softmax to compute certain measures for dissonance (\textsc{emd}) in our study.

% Counter({'ext': 0.6807009155108741, 'res': 0.4656857189337783, 'min': 0.28544508746002817})

% PL model to aggregate

%We only selected classes that all researchers agreed met our criteria. In addition we asked each annotator to estimate if this particular class would be easy to identify in an image. We marked images as difficult if at least one annotator indicated so. Based on this we ended up with 29 easy classes and 21 difficult classes.

\subsubsection{Crowdsourcing Task Design -- Image Recognition (Task-2)}
\label{sec:task-2}
Next, we aim to understand the factors that influence accuracy of humans in an image prediction task that is informed by the discriminative features identified by either other humans or by machines.  

In this task, workers were asked to identify an object in an image within a game-like experience. Workers were incrementally shown segments of an object in an image (one segment at a time), based on the aggregated human ordering (\human) or that corresponding to one of the 3 neural network models (\vgg{}, \inception{}, \resnet{}). In all cases, the segments were revealed according to a decreasing order of importance. The overall objective of the workers was to guess which object was being revealed, using as few uncovered segments as possible. The task began with one uncovered segment and workers could make at most 3 guesses by filling a text field after every new uncovered segment. Workers were also allowed to uncover another segment in case they did not have any guesses at each stage, by clicking a  `\textit{Show One More!}' button. To encourage workers to correctly identify the object using the fewest number of segments possible, we incentivized them with a bonus payment of 3 USD cents for every object they correctly identified using the fewest segments among the corresponding cohort of 5 workers for each image. After 50\% of an image was uncovered (i.e., 25 segments were shown), we automatically revealed the entire image and workers were allowed to make a final set of 3 guesses. We accepted misspelled guesses within an edit-distance of 1, and also expanded the list of acceptable responses by using a dictionary of synonyms. 
If workers failed to correctly identify the object, they were asked to identify whether the said object was present in the image using a multiple choice question (with `\textit{Yes}', `\textit{No}', or `\textit{I Don't Know}' options). Finally, all workers were then asked to respond to a question regarding how difficult it was to identify the given object in the image on a 5-point Likert scale ranging from `\textit{1: Very Easy}' to `\textit{5: Very Difficult}'. For convenience, we will refer to this task as Task-2 hereafter. 

%\ug{I will make another pass over this section once we decide on how to frame the motivation for our design; `feature matching' vs. `Biederman's theory'.}

\subsection{Measuring Image Difficulty and Dissonance}
\label{sec:diff-diss}

In this section we first introduce the notion of image difficulty and how it is computed in our setting. Recall that, in Task-2 (cf. Section~\ref{sec:task-2}), subjects are asked to assess the difficulty in identifying the object in the image (on a 5-point Likert scale) after the completion of their guessing procedure. In soliciting responses there is inherent variability in assessments of workers that might stem from factors such as their inherent familiarity with the object, sub-optimality of the segments being uncovered as a function of features choses by humans or machines, and so forth. 

\subsubsection{Difficulty of an Image} 
\label{sec:diff}

In coming up with an aggregate measure for inherent \emph{difficulty} of an image classification instance given a certain sequence of uncovered segments we assume the following:

\begin{itemize}
	\item We assume that for the same sequence of segments presented to humans (same model) there is inherently low variability in assessments.
	\item We assume that the optimal sequence for guessing, that is the best sequence that results in a successful guess in smallest number of segments, sets the difficulty of the task.
\end{itemize}

%We now ponder on these, seemingly limiting, 
We explore these assumptions and qualitatively argue their validity in guiding the design of our measure for difficulty. First, although there is variability in the number of uncovered segments needed to correctly guess the object in an image, we found low entropy in the self-reported difficulty assessments from corresponding workers. So for an image-model pair we take the median of the difficulty assessment values say $m_{i,j}$ where $i$ is the image and $j$ is the model that is generating the sequence (a neural network or humans). 

%Of course 
The optimal sequence of segments presented to the user that would solicit the best guess is unknown. We can however provide an upper bound to this by choosing the model that has the lowest difficulty estimate. Hence, we denote the \emph{difficulty} of an image $i$ as $\min_{j}\{ m_{i,j}\}$. 

Consider an image $i$ from our data set and NN $j$ (one of VGG, Inception or ResNet) with the number of segments needed to guess the correct label from 5 different crowd workers. For example we have the values (4,5,6,10,11). We now take the median of these assessments to get $m_{i,j}$ = 6. We compute this for each j in {VGG, Inception or ResNet} for the image i. Let's say these values are (6,10,21). Then the inherent difficulty of image i is the min of 6,10,21 which is 6. This gives us a data-driven measure of difficulty per image.

Note that this is different from the class level difficulty we solicited in the beginning of our experiments.
\subsubsection{Dissonance}
\label{sec:diss} 

Within the scope of our study, we propose two distinct notions of disparity between humans and machines (ML models); \textit{implicit} and \textit{explicit} \textit{dissonance}. We characterise \textit{implicit dissonance} as the difference between humans and machines emerging from the Task 1, due to collective differences in features (segments in our case) that humans and machines perceive as being more important for accurate classification. This plays a pivotal role in enabling workers to readily recognise images in the second task. We characterise \textit{explicit dissonance} based on the performance of humans and machines in Task 2. 

%The dissonance between humans and machines (ML models) can be \emph{implicit} or pertaining to the selection stage. In other words dissonance arising from differences in features (segments in this case) that humans and machines focus on or perceive as relevant for maximum success in the image recognition task. The other source of dissonance between machines and humans is more \emph{explicit} that compares the performance of humans and machines based on the success in the classification task. Note that the recognition task here refers here to Task 2 (cf. Section~\ref{sec:task-2}) where a human is trying to recognise the object in question. 

\mpara{Features.} Note that we used the same pixel clustering approach as that of SHAP when gathering judgements in Task-1 so as to ensure that the SHAP explanation is comparable to the data we gathered. Since the output of SHAP is an importance score (shapley value) distribution over segments, we order segments in decreasing order of these scores.
%\ug{The first sentence should be clarified further IMO. Explicitly describe the comparison. -- Addressed JS}

\mpara{Implicit Dissonance.} We quantify the \textit{dissonance} between human and machine understanding of these images as the distance between the human annotated segments and the output explanation of SHAP for each neural network model. We analysed the performance of the three neural networks with 3 measures having different semantics: Jaccard Similarity, NDCG~\cite{jarvelin2002cumulated}, weighted Kendall's $\tau$~\cite{shieh1998weighted} and EMD~\cite{rubner1998metric}. The simplest measure is coverage using Jaccard similarity between the human and machine annotated segments. Jaccard similarity however, does not capture the importance of segments indicated by their order of selection in our case. We use a weighted version of Kendall's $\tau$ to measure rank correlation between human and machine selection. Weighting here allows us to pay more attention to the ordering of the top segments. 
%NDCG is a metric often used to measure the performance of ranking algorithms. 

 $\tau$ entails order preservation but fails to capture locality. Locality is important because minor rank differences between segments that are spatially very close may be negligible. Earth Movers Distance (EMD)~\cite{monge1781memoire} is a Wasserstein metric that measures the distance between 2 distributions and takes locality into account. EMD between two sets of points in $\mathbb{R}^d$ of equal sizes (say, $s$) is defined to be the cost of the minimum cost bipartite matching between the two point sets. It is a natural metric for comparing sets of geometric features of objects. The EMD is based on a solution to the transportation problem from linear optimisation, for which efficient algorithms are available, and also allows naturally for partial matching. It is more robust than histogram matching techniques, in that it can operate on variable-length representations of the distributions that avoid quantization and other binning problems typical of histograms.

\mpara{Explicit Dissonance.} To get a more explicit notion of dissonance we use the data from Task-2. For each image $i$ we have the median number of segments needed to correctly classify it. Let $m_{i,j}$ denote the median number of segments needed to guess an image $i$ given model $j$'s segment ordering. We define \dissonance{} between a pair of models $j,k$ for $N$ images as the average difference in segments needed to correctly classify images.

%In order to aggregate those judgements to elicit the performance of guessing on this image, we take the median of count of segments revealed towards correct guess across all judgements, or $m_{i,k}$. %For a sequence of segments from model $k$ selected for image $i$ to be presented to evaluators we take the median of the \emph{number of segments guessed} from the multiple judgements obtained or $m_{i,k}$. 
%\ug{Rephrase this first sentence to make it more clear? - Addressed JS} 

$$
\dissonance{}(j,k) =  \left( \mathlarger{\sum}_{i} \frac{1}{Z} \| \{ m_{i,j} - m_{i,k}\} \| \right) / N  
$$

where $Z$ is the normalising factor that is chosen to be the maximum number of segments to bound the value between $[0,1]$. Intuitively, two models that differ to a large extent in the number of segments needed to guess can be safely assumed to be dissonant. 
Note that here the aggregated human ordering of segments can also be considered as a model $j$. 

% \subsubsection{Quantifying Dissonance Between and Difficulty of Human and Machine Understanding}To compare human and machine intuition we first need to identify reasons that the machine makes its predictions in the same format. Several posthoc interpretability methods have been suggested for models trained for computer vision related task. Many of these approaches can be used to identify either pixels or pixel clusters that the model considers important for a decision. In this work we utilise SHAP (and LIME?) to generate posthoc explanations  since it also estimates which pixel clusters are important. 

% explain SHAP in 4 lines

%Quantifying dissonance now is a matter of choosing an appropriate metric to calculate distance between the human annotated segments and the output of SHAP for each NN. We analysed the performance of 5 neural networks with 4 measures with different semantics: Jaccard Similarity, NDCG~\cite{jarvelin2002cumulated}, weighted Kendal\'s tau~\cite{shieh1998weighted} and EMD~\cite{rubner1998metric}.

% explain how each metric is computed briefly. one line and a formula will do.

% results discussion about which metric is good with qualitative examples. first explain what each measure captures and then systematically highlight its deficiencies. End with EMD being the most complete metric for this task

\section{Results}
\label{sec:results}
 
%In Study I we explored and quantified the difference between ML models and humans in the task of image classification. We found that among the best performing NNs, XXX is closest to human intuition. In Study II, we investigate whether the most discriminative features identified by ML models are superior to those identified by humans for the task of image classification. We delve into whether or not humans are best suited to identify features that fellow humans readily identify and understand.

\subsection{Human vs. Machine Ordering of Segments}
\label{sec:selection}
 
 By analysing the data gathered from our first task, we aim to understand how close the feature selection of machines is to human understanding (\textbf{RQ\#1}). %(cf. Figure~\ref{fig:num_seg_correct} and Table~\ref{tab:confusion}, ~\ref{tab:implicit_measures}). 

Human understanding is encoded in the segments selected by crowd workers and is operationalized by aggregations of these assessments from Task-1. % (cf. Section~\ref{sec:agreement-aggregation}).
It can be represented as a set (for precision), sequence or ordered list (for Kendall's $\tau$) or a distribution (EMD). Table~\ref{tab:implicit_measures} presents the differences in the implicit dissonance measures between humans and machines. We can clearly see that \inception{} and \resnet{} are closer to human intuition than \vgg{}. Using multiple one-way ANOVAs we found statistically significant differences between all the implicit metrics for dissonance across the three NN models at $p<.001$. We observe here that all of the measures are correlated to each other. Interestingly, we also see that the performance of the machine learned models on the official ImageNet test set (last column titled \textbf{Top-1 Acc}) are also correlated with human understanding. This finding relates to prior works, which have argued that models that correlate more with human feature selection tend to generalise better \cite{doumas2018human,geirhos2018generalisation}.%\ug{Add citations.}.
% This finding relates to prior works, which have argued that models that correlate more with human feature selection tend to generalize better \cite{}. \ug{Add citation.} 
% But does it mean that human selection also results in better performance in Task 2 ? We answer this question next.

\begin{table}[!ht]
\centering
\caption{Implicit Dissonance Measures -- How close are machines to human understanding when selecting features?}
\label{tab:implicit_measures}
\scalebox{.9}
{\begin{tabular}{lccccccccc}
\toprule
% 	p_5	p_10	emd	tau_c
% diff_val_bins2				
% 0	0.429939	0.446871	2.255686	-0.010328
% 1	0.449074	0.460185	2.302762	-0.000205
    & \multicolumn{2}{c}{\textsc{p@5}} & \multicolumn{2}{c}{\textsc{p@10}}  & \multicolumn{2}{c}{\textsc{emd}} & \multicolumn{2}{c}{\textsc{tau}} & \multicolumn{1}{c}{\textsf{Top-1 Acc}}   \\ 
    & easy & diff & easy & diff  & easy & diff & easy & diff &   \\ 
\midrule
\textbf{\inception}   & 0.89 & 0.83 & 0.77 & 0.71 & 1.5 & 1.9 & 0.31 & 0.23  & 80  \\ %\cline{1-1}
\textbf{\resnet}   & 0.87 & 0.81 & 0.75 & 0.69 & 1.6 & 1.9 & 0.30 & 0.20&  79 \\ %\cline{1-1}
\textbf{\vgg}  & 0.43 & 0.44 & 0.45 & 0.46 & 2.2 & 2.3 & 0.01 &  0.00 & 72 \\ %\cline{1-1}

% {\begin{tabular}{lcccccccc}
% \toprule

%     & \textbf{p5} & \textbf{p10} & \textbf{p20} & \textbf{kl} & \textbf{emd} & \textbf{tau} & \textbf{recall} & \textbf{acc} \\ 
% \midrule
% \textbf{\inception}  &  0.87  &   0.75  &   0.62  &   7.17 &  1.70 & 0.29 &   0.75     &     80\\ %\cline{1-1}
% \textbf{\resnet}  &  0.86  &   0.74  &   0.61  &  7.28  &  1.72   &    0.28    &  0.73 &  79\\ %\cline{1-1}
% \textbf{\vgg} &  0.43  &  0.45   &  0.47   &  10.90  & 2.32   & -0.01 & 0.98  & 72\\ %\cline{1-1}
\bottomrule
\end{tabular}}
\end{table}

Next, we explore whether NN models (\inception{}, \resnet{}, and \vgg{} in our case) which are closer to human intuition result in superior performance in the image recognition task. We are interested to see whether the sequence of segments aggregated from the segments selected by humans in Task-1, indeed result in better image recognition by other human subjects in Task-2.

\begin{figure*}[h]
 	\begin{subfigure}[b]{0.33\textwidth}
    \includegraphics[width=1\textwidth]{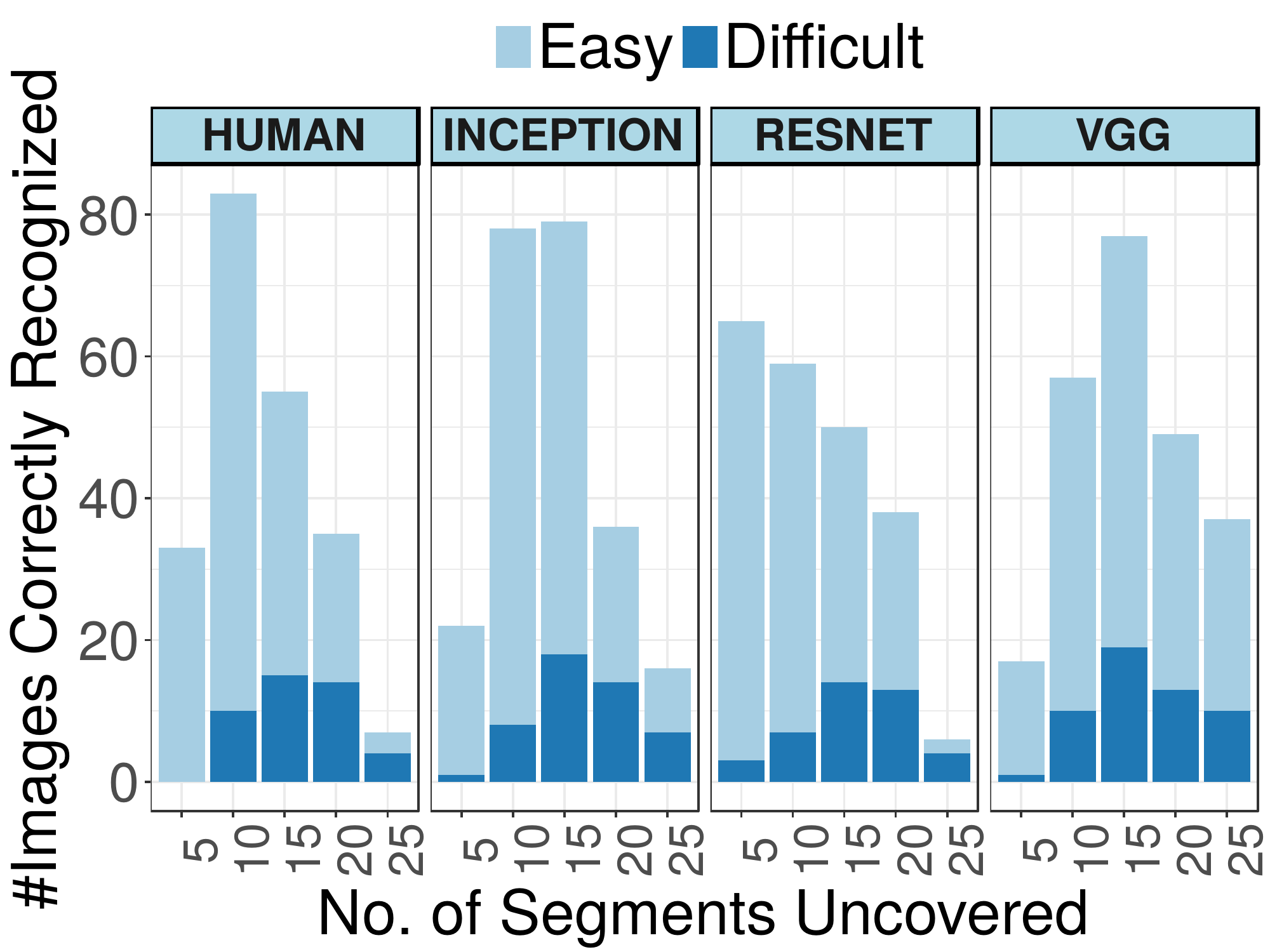}
    \caption{Humans versus Machines}
	\label{fig:segs_stacked_diff_all}
    \end{subfigure}
    %\begin{subfigure}[b]{0.24\textwidth}
    %\includegraphics[width=1\textwidth]{Img/segs_all}
    %\caption{}
	%\label{fig:segs_all}
    %\end{subfigure}
    ~
    \begin{subfigure}[b]{0.33\textwidth}
    \includegraphics[width=1\textwidth]{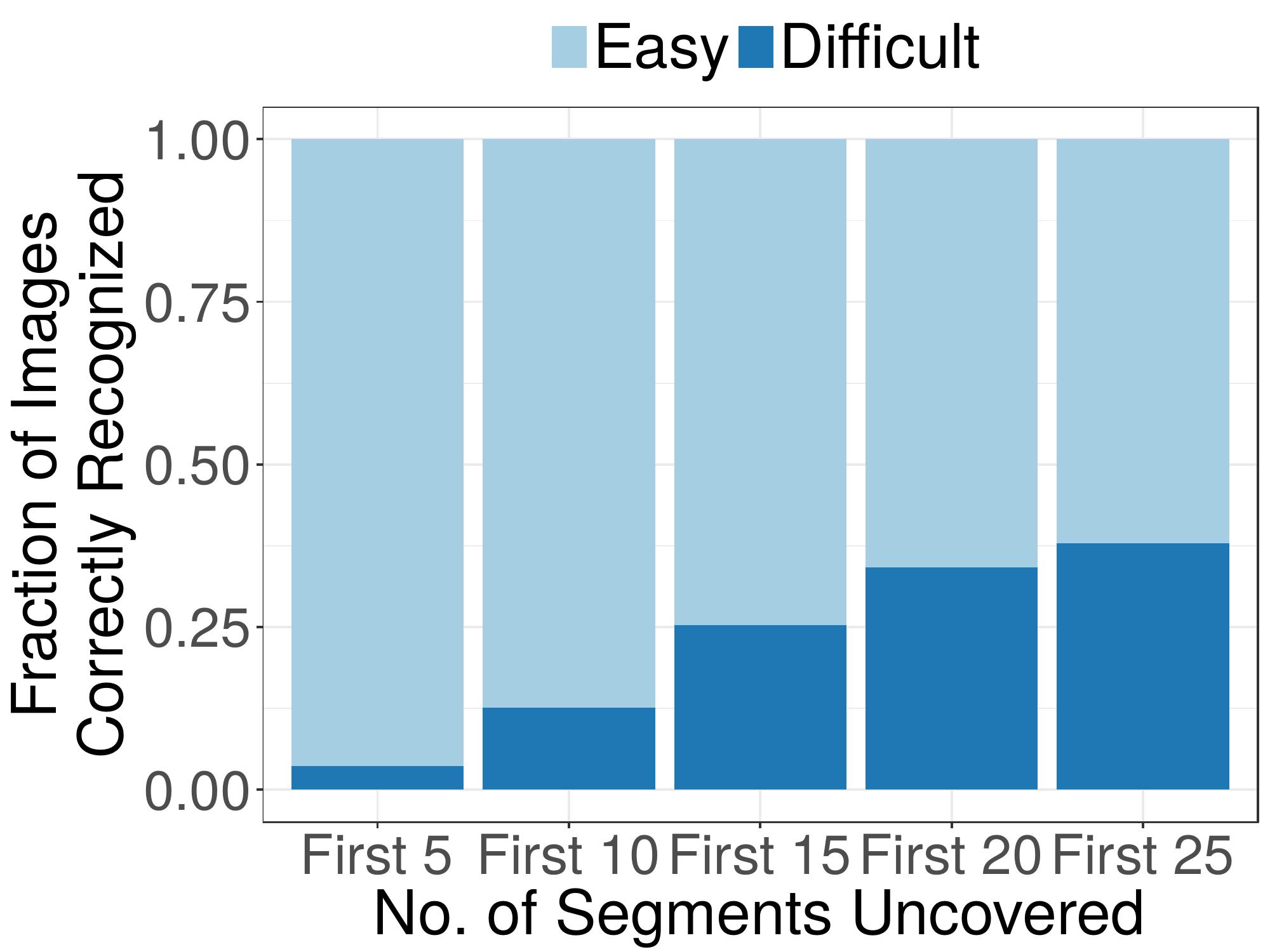}
    \caption{Humans and Machines} %(aggregated)}
	\label{fig:segs_all_diff}
    \end{subfigure}
    ~
	\label{fig:segs_all_results}
	\begin{subfigure}[b]{0.33\textwidth}
    \includegraphics[width=1\textwidth]{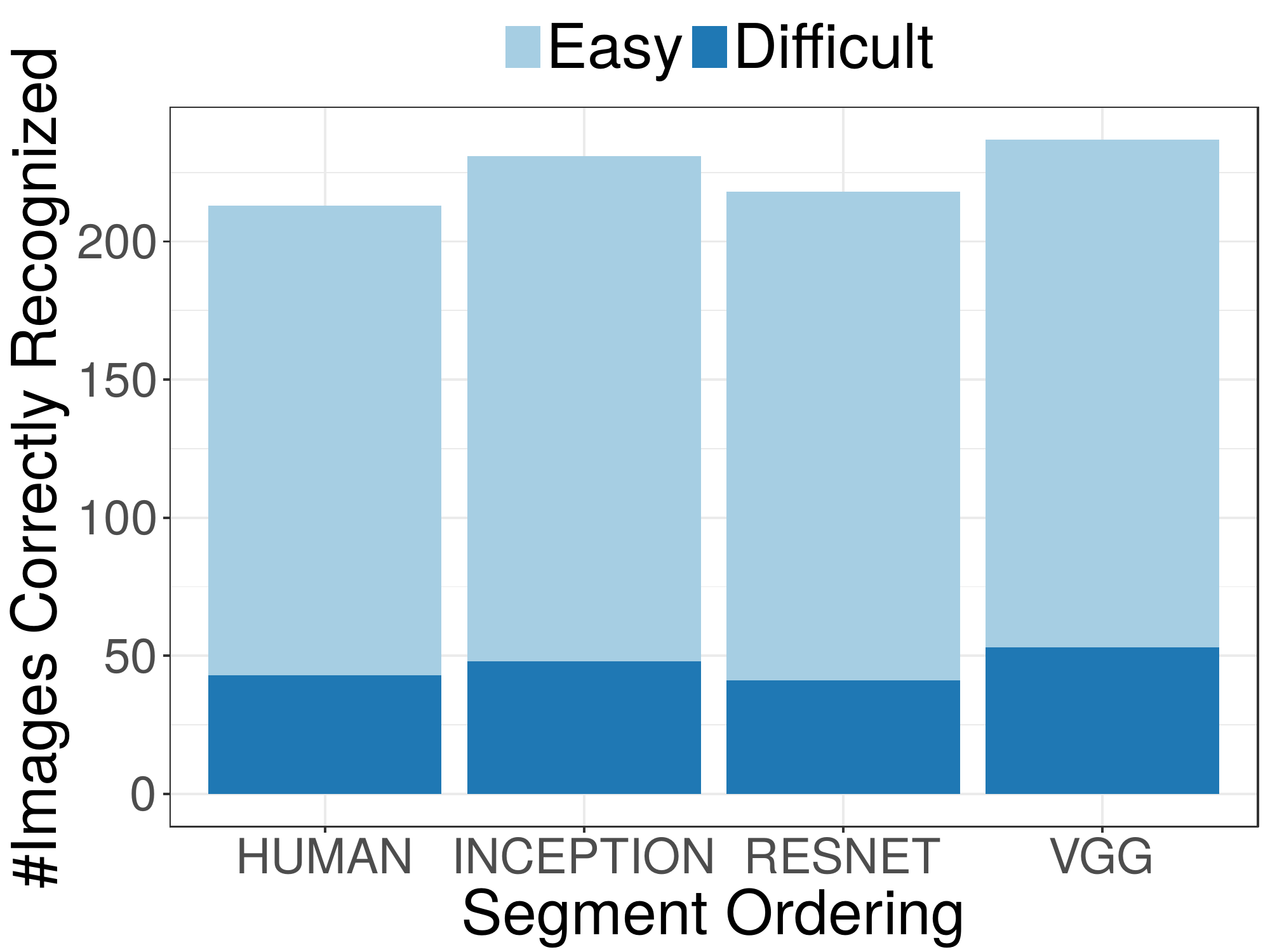}
    \caption{Humans versus Machines}
	\label{fig:segs_all_stacked}
    \end{subfigure}
    ~
	%\label{fig:segs_all_results}
	%\begin{subfigure}[b]{0.33\textwidth}
    %\includegraphics[width=1\textwidth]{Img/segs_stacked_diff}
    %\caption{}
	%\label{fig:segs_stacked_diff}
    %\end{subfigure}
    \caption{Distribution of easy and difficult images that were correctly recognised by workers in the guessing task (Task-2), where segments were uncovered in orders determined by humans ({\human{}}) in comparison to different machines (\inception{}, \resnet{}, \vgg{}).}
	\label{fig:segs_all_results}
\end{figure*}

Figure \ref{fig:segs_all_results} illustrates our findings. Contrary to what was expected, we found that human selection of important segments (\human{}) does not always lead to the best prediction by other humans. For the sake of readability, we present and discuss our findings in 5 segment intervals with respect to the number of segments uncovered for accurate image recognition. 
\resnet{} ordering resulted in the best performance by far in the image recognition task within the first 5 uncovered segments (62 images accurately recognised), when compared to \human{} (33 images accurately recognised), \inception{} (21 images accurately recognised) and \vgg{} (16 images accurately recognised)  as illustrated in Figure \ref{fig:segs_stacked_diff_all}. Note that our findings are consistent when the data is anlaysed in a continuous fashion without intervals. This is the first evidence which suggests that human understanding of feature selection is not the most discriminative for recognising images. 

\begin{figure}[h]
 	\begin{subfigure}[b]{0.3\textwidth}
    \includegraphics[width=1\textwidth]{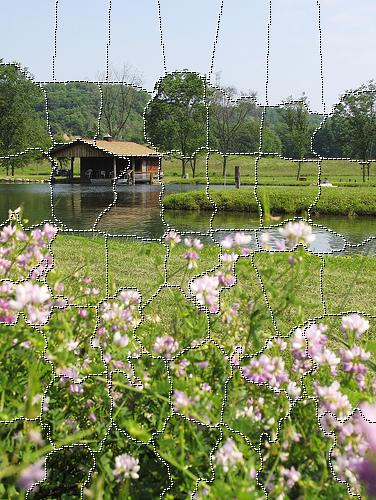}
    \caption{Original segmented image.}
	\label{fig:boathouse_segmented}
    \end{subfigure}
    ~\\
 	\begin{subfigure}[b]{0.23\textwidth}
    \includegraphics[width=1\textwidth]{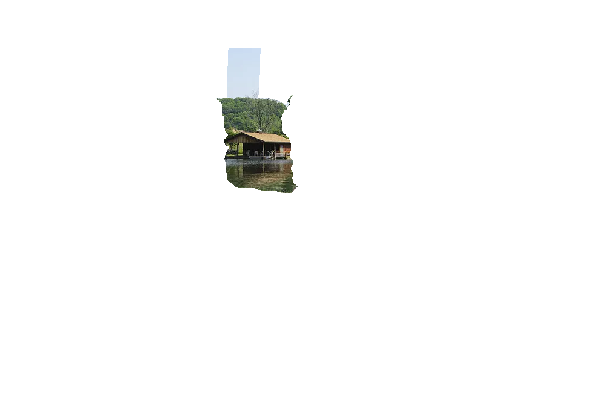}
    \caption{\human{}}
	\label{fig:boathouse_human}
    \end{subfigure}
    %~
    \begin{subfigure}[b]{0.23\textwidth}
    \includegraphics[width=1\textwidth]{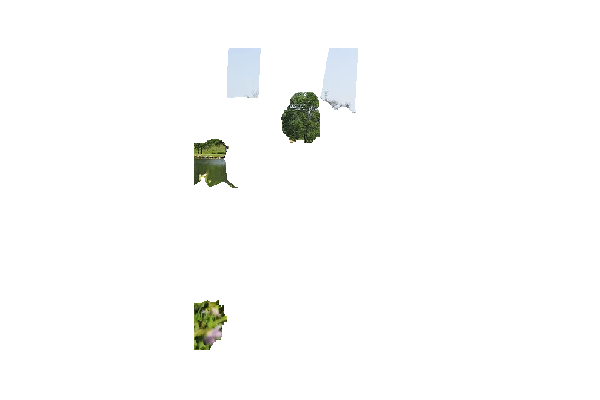}
    \caption{\inception{}}
	\label{fig:boathouse_inception}
    \end{subfigure}
    %~\\
	\begin{subfigure}[b]{0.23\textwidth}
    \includegraphics[width=1\textwidth]{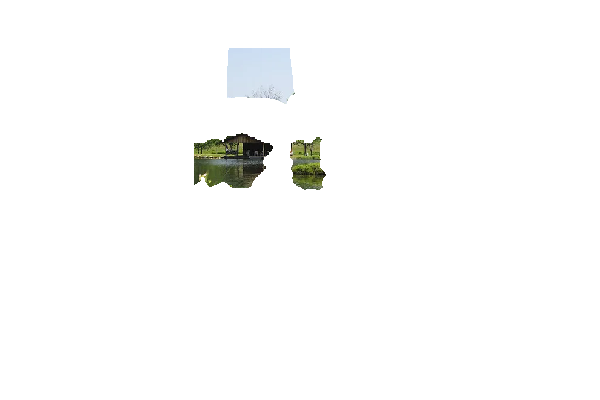}
    \caption{\resnet{}}
	\label{fig:boathouse_resnet}
    \end{subfigure}
    %~
    \begin{subfigure}[b]{0.23\textwidth}
    \includegraphics[width=1\textwidth]{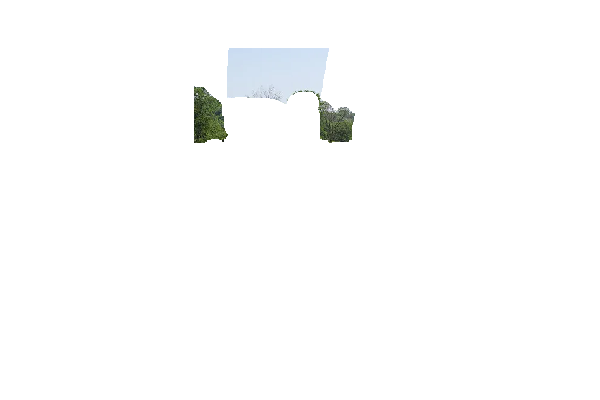}
    \caption{\vgg{}}
	\label{fig:boathouse_vgg}
    \end{subfigure}
    \caption{An example of a segmented image from 
    the `\texttt{boathouse}' class (\ref{fig:boathouse_segmented}) as displayed to humans in Task-1, and 5 of the most discriminative segments uncovered in Task-2 (\ref{fig:boathouse_human}, \ref{fig:boathouse_inception}, \ref{fig:boathouse_resnet}, \ref{fig:boathouse_vgg}), where humans (\human{}) selected segments covering the boathouse mostly in the first 5 selections, while machines (\inception{}, \resnet{}, \vgg{}) tend to select contextual segments including the sky, river, and grass.}
	\label{fig:human_better}
\end{figure}

Image recognition based on \human{} ordering catches up with \resnet{} as more segments are uncovered. Figure~\ref{fig:human_better} shows an example image where humans were able to better identify discriminatory segments. Human selection is independent of the dataset biases that the NNs are exposed to during training.

Interestingly, we found that segment ordering based on \inception{} and \vgg{} results in an increase in the number of images correctly recognised by humans in Task-2 after the uncovering of around 10 segments. %\ug{Could this be due to the depth of the NNs? Resnet vs. VGG and Inception.}
%Surprisingly \inception{} and \vgg{} are also lower in dissonance as compared

\textbf{Why \resnet{} why}: We further examined the images where \resnet{} performed considerably better than humans in Task-2. We consistently found that while humans particularly focus on the segments belonging to the given object, \resnet{} and the other machines in general, also focused on discriminative features outside the body of the object that comprise the context. This is illustrated in Figure \ref{fig:kimono-example}. We see that \inception{} and \resnet{} also pick the faces of the women which is rich context for guessing the correct label of the image, `\texttt{kimono}'.
%resulted in higher guess accuracy.

The importance of context for image recognition is well documented in human cognition literature \cite{auckland2007nontarget} as well as machine learning \cite{lawson2014leveraging}. %This is our second important finding
Thus, we reveal that although humans are good at classifying images, they do not always perform well in selecting the most discriminatory features for image recognition in our setting. It is indeed the case that we do not explicitly ask crowd workers to select discriminative segments with respect to the nature of task 2 but neither are the NNs trained specifically to help humans determine the class label in the fewest segments.

In fact we found that \human{} ordering helped other humans guess the fewest images overall (217). \vgg{} helps users guess the most images correctly (244) albeit slowly (i.e., after several segments are uncovered). We reason that since \vgg{} tends to overfit and memorize more patterns, it is able to eventually present good enough segments to facilitate a correct answer for most images. Our findings suggest that deeper networks with residual connections like \resnet{} learn similar abstractions for image understanding as humans and hence are capable of identifying the segments most essential for accurate image recognition.
%even with a small number of segments are able to focus on segments most essential for a human. 

\resnet{} has highest explicit dissonance (0.221) while also helping humans guess the most objects within the first 5 segments. Interestingly, in this light, it reinforces the finding that \resnet{} selects more discriminatory features early on compared to \human{}. \inception{} (0.209) and \vgg{} (0.207) are less dissonant but do worse than humans in estimating the importance of discriminative features. \vgg{} exhibits negative $\tau$ but still gets the most number of correct guesses overall by revealing key object segments after the context segments. \vgg{} also exhibits high EMD indicating its tendency to cater to context first.

\begin{figure*}
 	\begin{subfigure}[b]{0.33\textwidth}
     \includegraphics[width=1\textwidth]{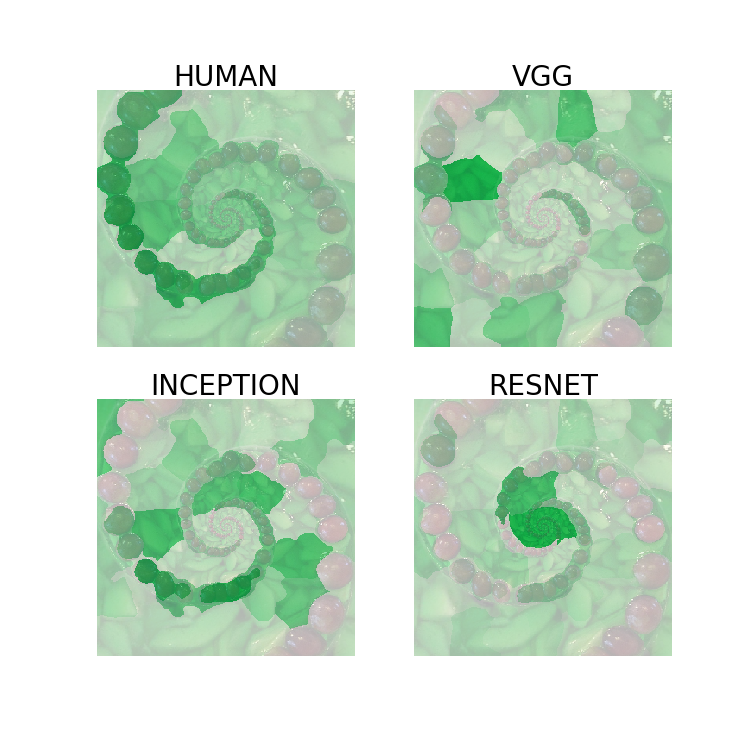}
    \caption{\texttt{coil}}
	\label{fig:coil}
    \end{subfigure}
    ~
    	\begin{subfigure}[b]{0.33\textwidth}
    \includegraphics[width=1\textwidth]{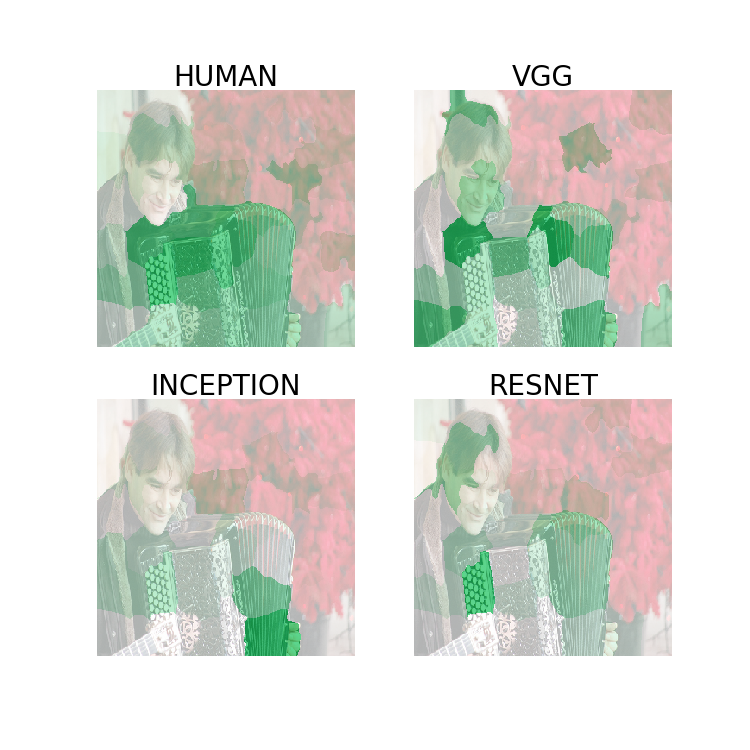}
    \caption{ \texttt{accordion}}
    \label{fig:accordion}
    \end{subfigure}
    ~
	\begin{subfigure}[b]{0.33\textwidth}
    \includegraphics[width=1\textwidth]{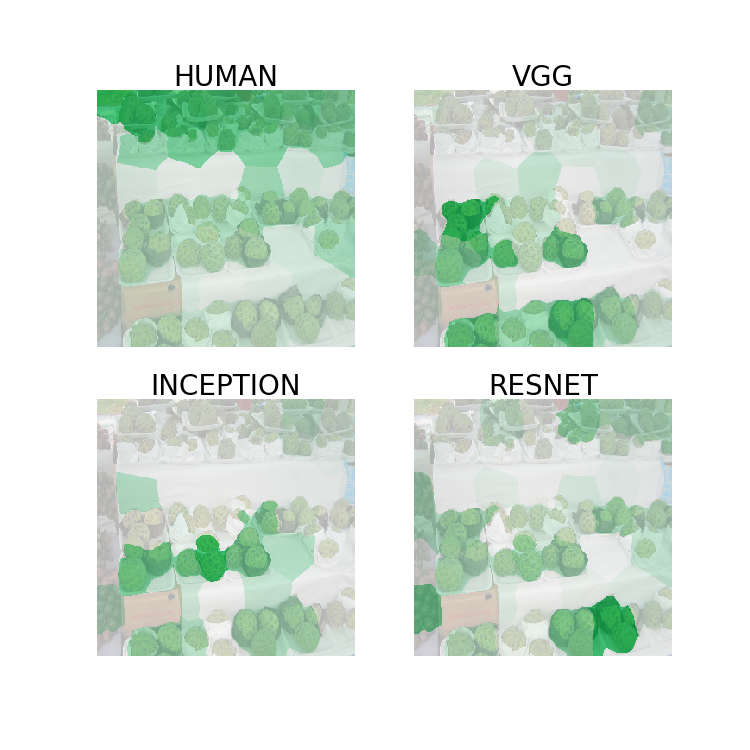}
    \caption{ \texttt{custard apple}}
    \label{fig:capple}
 	\end{subfigure}
    ~
    \\
    \begin{subfigure}[b]{0.33\textwidth}
      \includegraphics[width=\textwidth]{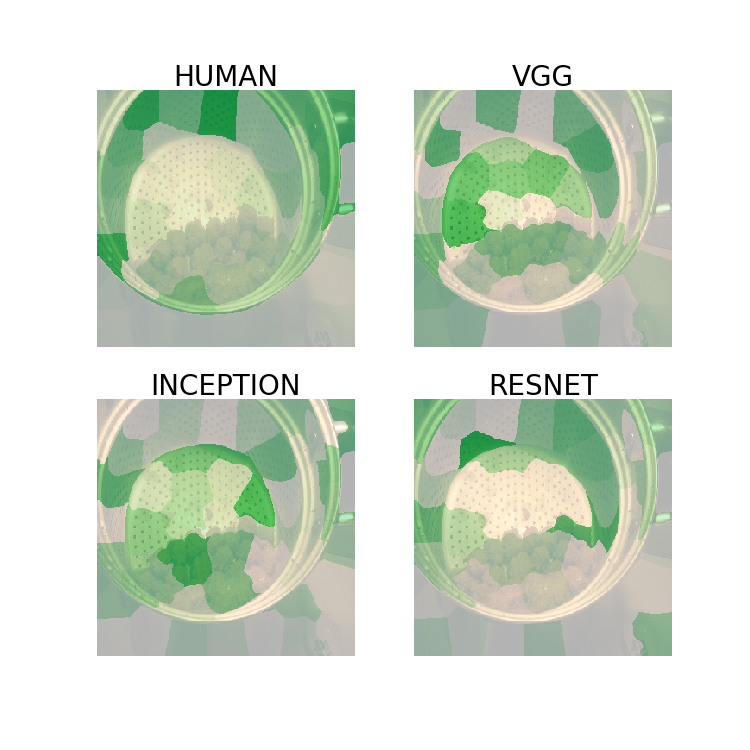}
    \caption{\texttt{strainer}}
    \label{fig:strainer}
 	\end{subfigure}
 	~
 	\begin{subfigure}[b]{0.33\textwidth}
      \includegraphics[width=\textwidth]{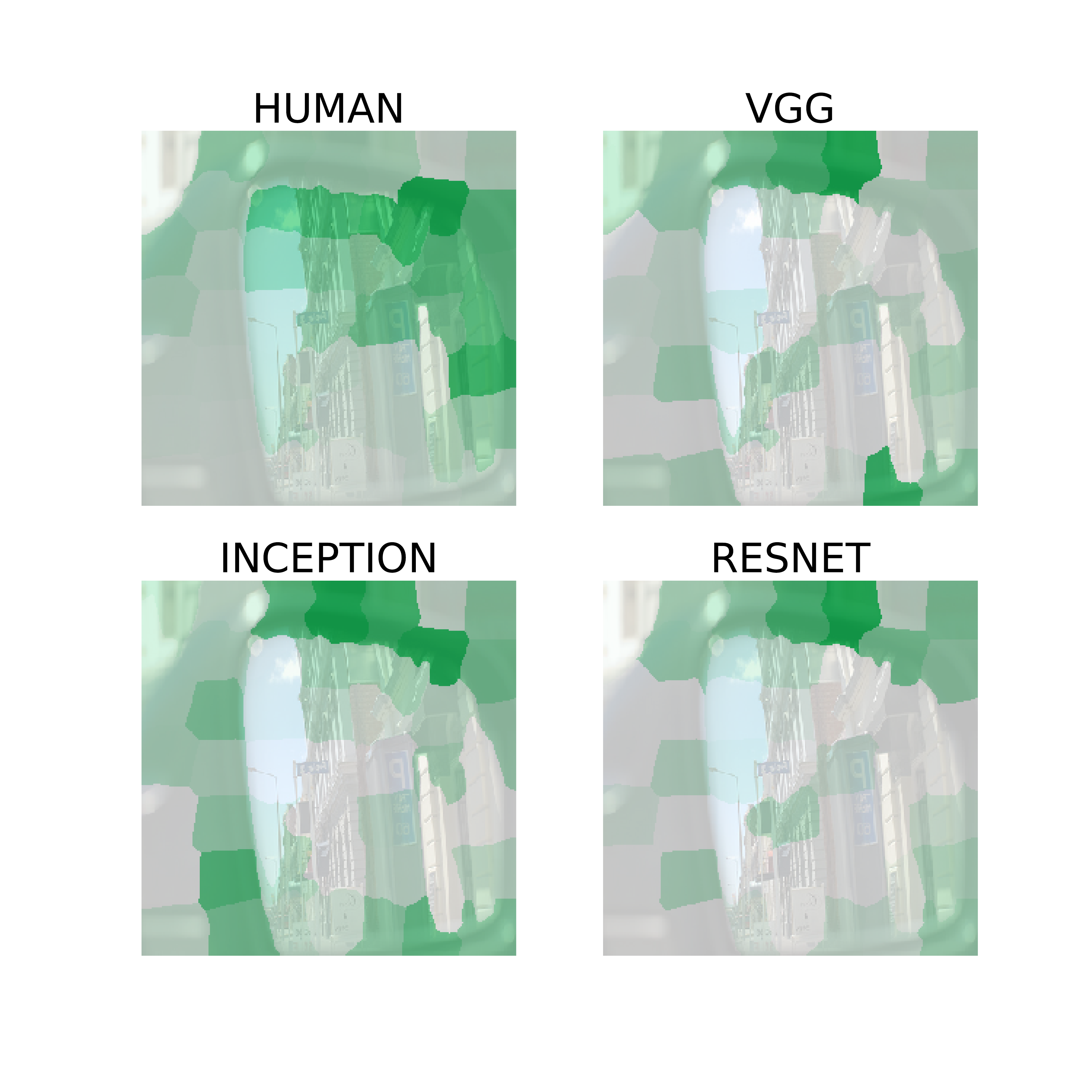}
    \caption{\texttt{car-mirror}}
    \label{fig:car-mirror}
 	\end{subfigure}
	
    \caption{Heat maps encoding the order of segment importance corresponding to humans and machines (in clockwise direction from top-left: \human{}, \vgg{}, \inception{}, \resnet{}.) The heat map is a visualisation of the importance scores returned by SHAP. The intensity of the green colour shows the relative importance between segments. In Task-2, the segments are shown in the order of intensity as displayed in these heat maps. Segments with no coloration have an importance score of 0 or lower. }
    %\ug{Describe how to read the heatmap. What do the colors mean? What does the intensity of the color mean?}
	\label{fig:examples}
\end{figure*}

Figure~\ref{fig:coil} illustrates how \resnet{} selects the best feature to guess \texttt{coil} and has median number-of-segments-to-correct-guess of 2 while all others require more than 20. This is in accordance with Biederman's \textit{recognition-by-components theory}, where he showed that a delay in the determination of an object's components has an effect on the identification latency of the object \cite{biederman1985human}.

%\js{i feel like we need more of an explanation here as to why human ordering leads to lowest amount of images guessed. we theorized that it was because in task 1 people didn't select more than 15 segments or so. this meant that after 15 we were showing random.} \ug{Yes, but this would reflect poorly on our task design. Retrospectively that's probably what casued it though. Is there a way to provide an explanation without painting a target for concerns?}

\begin{framed}
Therefore, with respect to \textbf{RQ\#1} we found that humans are not always superior to machines in selecting discriminative segments in images. \resnet{} ordering led to the most number of correct guesses within the first 5 segments.
\end{framed}

% TODO: definitely need an example here

%This artifact results in . \human on the other hand is the lowest with 217. 
%In contrast, the number of correctly guessed images on \human{} ordering monotonically decreases after 10 segments are uncovered. 

%One reason to explain this phenomena is the fact that we do not gather human ordering judgments for every segment in an image. Once we run out of human annotations the segment ordering is uniform which could lead to low information gain. This quirk is common to \inception{} and \resnet{} as well since they are deep networks with fewer parameters and focus on only the most important segments.

% The cell $(i,j)$ corresponds to the number of cases (number of images) when $i$ outperforms $j$. Success is measured by when a guessing sequence indeed resulted in correct identification (top values).  To measure performance we use \emph{relative success rate}, i.e., number of instances when model $i$ had more successes than $j$ in each of its 5 assessments. On the lower part of the table we report the \emph{domination} values.

\subsection{Effect of Image Difficulty on Human and Machine Understanding }
\label{sec:effect_of_diff}
%\subsection{Image Difficulty and Dissonance Between Human and Machine Understanding}
%We studied this impact of image difficulty, across the segment ordering resulting from the aggregation of human selections, in juxtaposition with machines. 
In this section we elaborate further on the impact of image difficulty in both, segment selection (Task-1) and object recognition (Task-2). 

\textbf{Task-1}: On analyzing the segments selected by humans and machines, we found that the average number of segments selected by humans and different NNs is nearly the same ({\raise.17ex\hbox{$\scriptstyle\mathtt{\sim}$}}18 for easy images, {\raise.17ex\hbox{$\scriptstyle\mathtt{\sim}$}}17 for difficult images, as shown in Table \ref{tab:task_difficulty}). For the number of segments selected by a NN we only considered segments with a positive score as returned by SHAP. Segments with a positive score are those which directly contribute towards the correct classification decision. 
%\ug{Explain this in a sentence or two (CSCW audience).}.

However, on average across all segment orders, humans successfully recognize objects in Task-2 after uncovering around 10 segments of the easy images and 15 segments of difficult images. We conducted two one-way between subjects ANOVAs to investigate the effect of image difficulty (for each \textit{easy} and \textit{difficult}) on the average number of segments uncovered to elicit accurate image recognition across the segment ordering conditions (\human{}, \inception{}, \vgg{}, \resnet{}). In case of the \textit{easy} images, we found a significant difference across all conditions; $F(2,618)= 39.22$, $p<.001$. Similarly, in case of the \textit{difficult} images, we found a significant difference across all conditions; $F(2,276)= 3.75$, $p<.05$. Post-hoc Tukey HSD tests revealed a significant difference between \resnet{} and the other three models ($p<.001$) in case of \textit{easy} images, while it revealed a significant difference between \resnet{} with respect to each of \inception{} and \vgg{} ($p<.01$) in case of \textit{difficult} images.  Thus, we found that \resnet{} needs the least uncovered segments for successful object recognition in case of both \textit{easy} (8.7 segments) and \textit{difficult} images (14.3 segments) on average, followed by \human{} with 9.6 segments for \textit{easy} and 14.6 segments for \textit{difficult} images. %This reiterates that \resnet{} and \human{} are effective in identifying the important segments of the image. 

A two-tailed T-test revealed a significant difference in the the average number of segments uncovered to elicit accurate image recognition in Task-2 based on the image difficulty (\textit{easy, difficult}), across all models (humans and machines in aggregate); $t(2,898)=36.90$, $p<.001$. This supports our intuition that \textit{easy} images can be recognised more quickly than the \textit{difficult} counterparts.

\begin{table*}[!ht]
\centering
\caption{Comparison of the number of discriminative segments selected in Task-1 and the number of segments uncovered before eliciting accurate image recognition in Task-2, across different models (humans and machines) and with respect to \emph{inherent} image difficulty (easy, difficult).}
\label{tab:task_difficulty}
\scalebox{0.75}{\begin{tabular}{lrrcrrcrrcrrc}
\toprule
    & \multicolumn{2}{c}{\human{}} & &\multicolumn{2}{c}{\inception{}}  & & \multicolumn{2}{c}{\vgg{}} & & \multicolumn{2}{c}{\resnet{}}\\ 
    \cline{2-3} \cline{5-6} \cline{8-9} \cline{11-12}     
    & \textbf{Easy} & \textbf{Difficult} & & \textbf{Easy} & \textbf{Difficult} & & \textbf{Easy} & \textbf{Difficult} & & \textbf{Easy} & \textbf{Difficult} \\ 
\midrule

\textbf{\#Segments Selected (Task-1, avg.)} & 17.9 & 17 & & 17.9 & 17.2 & & 17.9 & 17.1 & & 17.9 & 17.0 \\
\textbf{\#Segments Uncovered (Task-2, avg.)} & 9.6 & 14.6 & & 10.7 & 15.5 & & 13.1 & 15.8 & & 8.7 & 14.3 \\
\textbf{\#Segments Uncovered (Task-2, median)} & 8.6 & 13.9 & & 10.2 & 15.4 & & 12.5 & 15.0 & & 8.0 & 15.0 \\
\textbf{Time Taken (Task-2, in seconds, avg.)} & 119.6 & 175.9 & & 85.5 & 123.7 & & 90.8 & 113.3 & & 82.2 & 110.7 \\
\textbf{\#Classes}  &  44 & 30& & 46 & 36 & & 46 & 34 & & 47 & 29 \\ 
\textbf{\#Images}  &  172 & 45& & 183 & 53 & & 186 & 58 & & 179 & 43 \\  
 
\bottomrule
\end{tabular}}
\end{table*}

We conducted two one-way between subjects ANOVAs to investigate the effect of image difficulty (\textit{easy} and \textit{difficult}) on the average amount of time taken by human assessors in Task-2 to accurately recognize the images across the different segment ordering conditions (\human{}, \inception{}, \vgg{}, \resnet{}). In case of the \textit{easy} images, we found a significant difference across the conditions; $F(2,618)= 4.48$, $p<.05$.
Post-hoc Tukey HSD test revealed a significant difference between \human{} segment ordering with respect to all three neural networks, ($p<.001$). We did not find a significant effect across the conditions in case of the \textit{difficult} images. Our findings show that human assessors took more time to recognize images accurately when the segments were revealed according to \human{} ordering in comparison to each of the neural networks, when the images were \textit{easy}. We reason that this is because the neural networks focus on the context early on, whereas humans tend to select the whole object first which may still make it hard to identify the object without the aid of contextual cues.

%\ug{Can we add an explanation for this observation? Can we say the following? ``This follows our previous finding that humans are not necessarily better than machines in identifying the most discriminative segments in images. Hence, even in \textit{easy} images the HUMAN ordering tends to inhibit quick image recognition on average.''}

\textbf{Task-2}: We first explored the nature of image classes in our dataset with respect to the class membership of images that were correctly recognised. We define a class as being \textit{covered} if at least 5 of the 6 images from the class are correctly recognised in Task-2. We present the class coverage resulting from human and machine segment ordering in Table~\ref{tab:class_coverage}. We found that \vgg{} corresponds to the highest class coverage of 68\%, while \human{} corresponds to the lowest. 
%\js{A little explanation of why humans suck here would be nice.}

\begin{table}[h]
\centering
\caption{Class coverage resulting from segment ordering by humans and different machines. Bold classes are covered *only* by the corresponding model.}%The bold classes are even preferred by the corresponding model only - no other models can cover this class.}
\label{tab:class_coverage}
\scalebox{1}{
\begin{tabular}{lcl}
\\\toprule
\textbf{Ordering} & \textbf{Class Coverage} & \textbf{Example Covered Classes}\\
\hline
\textbf{\inception} & \texttt{60\%} &\texttt{strainer}, \texttt{water buffalo}, \texttt{scoreboard}, \texttt{...}\\
\textbf{\resnet} & \texttt{60\%} & \textbf{\texttt{dam}}, \textbf{\texttt{milk can}}, \texttt{cannon}, \texttt{...}\\
\textbf{\vgg} & \texttt{68\%} & \textbf{\texttt{freight car}}, \texttt{strainer}, \texttt{kimono}, \texttt{...}\\
\textbf{HUMAN} & \texttt{58\%} & \texttt{boathouse}, \texttt{common iguana}, \texttt{car mirror},  \texttt{...}
\\\bottomrule
\end{tabular}}
\end{table}

Across all images that were correctly recognised using human and NN ordering of segments, we observe the expected trend of easy images being recognised quickly (with fewer uncovered segments) and the difficult images requiring more uncovered segments before being correctly guessed (as shown in Figure \ref{fig:segs_all_diff}). Finally, we also found that the \vgg{} segment ordering was most effective in correctly recognising difficult images in comparison to \human{} and other machines (see Figure \ref{fig:segs_all_stacked}). 

In Table~\ref{tab:confusion},  we present a confusion matrix of cases when a given model (human or machine) performs better or dominates another. 
 
 \begin{table}[!h]
\centering
\caption{Confusion matrix of model domination. Domination values are counts in two image scenarios: Easy/Difficult.}
\label{tab:confusion}
% \small
% \scalebox{.9}{
\begin{tabular}{lccccc}
\toprule

% \textbf{Inception-V3} & - & (tbd) & 21easy, medium, hard & 38 \\
% \textbf{Resnet} & (tbd) & - & (tbd) & (tbd) \\
% \textbf{VGG} & 28 & (tbd) & - & 40 \\
% \textbf{Human} & 20 & (tbd) & 15 & - \\
% \textbf{Inception-V3}  &  -  &   12 (tbd) &   XXX  &   XXX \\ 
% \textbf{Resnet}  & XXX  &   -  &   XXX  &   XXX \\ 
% \textbf{VGG}  &  XXX  &   XXX  &   -  &   XXX \\ 
% \textbf{Human }  & XXX  &   XXX  &   XXX  &   - \\ 

% &&\textbf{Success Rate}&&\\
% \hline
%     & \textbf{\inception} & \textbf{\resnet} & \textbf{\vgg} & \textbf{Human } \\ 
% \midrule
% \textbf{\inception} & - & 18/20 & 10/11 & 18/20 \\
% \textbf{\resnet} & 15/13 & - & 13/9 & 22/10 \\
% \textbf{\vgg} & 13/15 & 21/20 & - & 18/22 \\
% \textbf{Human} & 8/12 & 17/12 & 5/10 & - \\
% \hline
% &&&&\\
% &&\textbf{Model Domination}&&\\
% \hline
    & \textbf{\inception} & \textbf{\resnet} & \textbf{\vgg} & \textbf{Human } \\ 
\midrule
\textbf{\inception} & - & 66/31 & 103/22 & 81/31 \\
\textbf{\resnet} & 118/29 & - & 124/23 & 104/26 \\
\textbf{\vgg} & 81/42 & 63/39 & - & 72/42 \\
\textbf{Human} & 101/25 & 73/25 & 109/18 & - 
\\\bottomrule
\end{tabular}
% }
\end{table}
 
A model is said to dominate another on a given instance if it takes, on average, a lesser number of segments for the worker to recognize an image. So cell $(i,j)$ is the count the number of instances when model $i$ dominates $j$ in terms of the number of segments required to guess the correct image type. We present domination values as counts in two scenarios -- when the image is considered to be \textbf{easy}, and \textbf{difficult}.

Consider the difference between \human{} selection and \resnet{}. \human{} selection dominates \resnet{} on 73 easy images, as opposed to being dominated on 104 easy images by \resnet{}, wherein \resnet{} segment ordering leads to correct recognition with fewer uncovered segments. 
%\js{any other insight this can give us into where the gaps are in human vs machine understanding?}

%This is the first evidence of the fact that human understanding of feature selection might not be the most discriminative for recognising images. 

 %This shows that humans 

\begin{framed}
Addressing \textbf{RQ\#2}, we found that image difficulty, the order and the number of discriminative segments revealed influence the accuracy of humans (i.e., crowd workers in Task-2) in the image recognition task.
\end{framed}

\section{Discussion}
\label{sec:discussion}

    \textit{Demographics of Participants --} To maintain the integrity of the experimental setup and not divert worker attention from the task at hand, we did not gather explicit background information from crowd workers regarding their demographics. Based on the data available by default from the Figure8 platform, we found that 68 distinct trustworthy workers from 17 different countries completed 1,500 instantiations of the image classification task (300 images $X$ 5 judgements). In Task-2, 309 distinct trustworthy workers from 41 different countries completed 6,000 instantiations of the image recognition task (300 images $X$ 4 segment ordering models $X$ 5 judgements). We did not find any significant influence of country of origin of workers on the characteristics of segments selected in Task-1 or the objects recognised in Task-2.

\textit{Key Takeaways --}  Our results revealed interesting insights into how both humans and machines approach the task of object recognition in images. The first key takeaway is that humans %may not be the best when it comes to selecting discriminative segments.
are not consistently better than machines when it comes to selecting discriminative segments in images. 
From our study we see that \resnet{} is better than \human{} in helping workers quickly identify images. \resnet{} is a deep neural network with residual connections that helps to better train a deep network. We see that deep networks (including \inception{}) select good discriminative features when compared to the denser and shorter \vgg{}. We ascertain these features to be discriminative due to the support from  Biederman's work on `\textit{human image understanding}', where he showed that a delay in the determination of an object's components leads to increased latency of the object recognition.
We note that some of the interesting scenarios where humans are worse than neural network models in selecting discriminative features for recognition, open up interesting avenues for future work. In particular, understanding how humans perceive context and the role that context plays in human understanding can be pivotal in building more human-like machines. Our work presents an important first step towards the vision of thoroughly understanding the dissonance between humans and machines across a variety of tasks.

% We hypothesize that deeper networks with residual connections like \resnet{} learn similar abstractions for image understanding as humans and hence even with a small number of segments are able to focus of segments most essential for a human. 

\subsection{Caveats and Limitations}

\textit{Crowdsourcing Setup --} We took several measures to ensure the reliability of responses gathered from crowd workers in Task-1 and Task-2 \cite{gadiraju2015understanding,gadiraju2017clarity}. By using dynamic worker lists, we made sure that workers participated in only one crowdsourcing task in our entire study. Workers in Task-1 were not allowed to participate in Task-2, and workers were not allowed to participate in more than one condition within Task-2 (\human{}, \inception{}, \resnet{} or \vgg{}). We chose not to show workers in Task-1 all 1000 ImageNet classes, since in our pilot study workers exhibited a tendency to repeatedly guess the label of images they were previously exposed to in the task, on encountering a new image to recognise. We accounted for this in our final study setup described in Task-1 by limiting the number of images being shown and ensuring that only images from distinct classes were shown to each worker. Showing workers all 1000 classes beforehand would also have potentially increased their cognitive load significantly, thereby biasing our experimental setup.

\textit{Recognition-by-Objects --} We adopt a simplified understanding of Biederman's theory \cite{biederman1985human} for object recognition. Note that in the original theory that was proposed, Biederman showed that a set of components could be derived from five properties of edges in a 2-dimensional image; curvature, co-linearity, symmetry, parallelism and co-termination. Since the detection of these properties has been shown to be invariant to the quality of the images and the viewing position, we project this notion of components onto image `segments' in our case. 

\textit{Super pixel segmentation --} By operating on this space for both humans and neural networks, we make comparison easier and more accurate. Using free form annotations from humans as an alternative to super pixel segmentation would make agreement computation complex, aggregation of annotations hard, and introduce noise in the metrics. 

\textit{Framing of Task-1 --} Our goal within Task-1 was to understand how humans select important segments for identifying the given object in an image. It was therefore important to frame the task without confounding it with an end goal of helping other humans recognise the object. Our rationale behind this is that in the image classification task, ML models also operate with an aim to correctly identify the object in the image. The aim of Task 2 was to then measure the impact of the dissonance between human and machine understanding of images where other humans are tasked with recognising an object being revealed one segment at a time. Further experiments are required to test whether framing the task differently, and asking workers to focus more on the segments they would pick to help other humans identify the object would have a significant impact on their segment selection process.

\textit{Selection of Discriminatory Segments --} It must be acknowledged that an alternative hypothesis that can explain the segment selection process of humans in Task-1 is the possibility that humans potentially use more context in their decision making process than they attribute to it. Another potential factor that may influence the segment selection process of humans, is the noise in their ranking of segments in the decreasing order of importance beyond the first few segments.

%Firstly, note that for \human{}, with our design of Task-1, it was difficult to estimate the relative ordering of all segments in an image without making the segments too big and uninformative. \ug{This first sentence isn't clear to me. Clarify?}

\textit{Choices Made for Segment Ordering --} Secondly, Using SHAP with deep models possessing fewer parameters (\inception{} and \resnet{}), only gives us the distribution of importance on the segments the network focuses on. Since they are smaller models they focus on fewer segments and we do not have the overall ordering of all 50 segments. For some images, it is also a difficult task to order all 50 segments in an image accurately even for humans. To overcome such cases, once we run out of annotations/importance estimates, the segment ordering corresponding to the rest of the image is uniformly random which could lead to low information gain. Using SHAP with \vgg{} on the other hand results in information about nearly all segments which could be another indicator as to why it corresponds to the most images accurately recognised overall in Task-2. 

\subsection{Implications for CSCW and HCI}
An important goal for CSCW and HCI research today is to make AI systems more receptive of human needs. Understanding human-machine dissonance (eg. through answering which neural network is more human like), has direct applications in evaluating and building credible and interpretable machine learning models \cite{wang2018learning} which can support and shape our everyday interactions. Users are more likely to trust and adopt credible models where explanations conform to established domain understanding. We provide metrics to understand dissonance, and a data set that the community can use for evaluation and training models for object recognition. Our work can inspire and inform further studies that evaluate the ``human-ness'' of neural network models in different tasks, both from a design choices standpoint and through our findings.  

\textbf{Ethical implications of our work.} Our study can inform and further the ethical discussions around machine learning models in terms of their congruence with human expectations. Machine learning models increasingly mediate our daily lives, nudging human behaviour along the way \cite{rahwan2019machine}. However, with the boon of nudging human behaviour in a positive direction or intended way comes the risk that human behaviour may be nudged in undesirable or unintended ways. For example, people can be influenced to buy certain products, or watch particular television programs, or even vote for particular political parties.

We aim to better understand the congruence of human expectations with machines by studying the dissonance between humans and machines. We use the lens of the image classification task, analysing segments of the image that humans consider as being important to classify a given object in contrast to machines. Images where humans take longer to determine the class label (a higher number of segments in Task 2) or justify their decisions differently (according to metrics like EMT and tau in Task 1) as compared to a neural network model are strong indicators of human and machine misalignment. Understanding such disagreement can help to reason about whether the misalignment is ethically sensitive, i.e. is the model making the right choice for ethically or morally wrong reasons. Secondly, since Task 2 does not explicitly inform subjects about the source of the segment ordering (humans or machines), we can potentially further analyse which neural network models imbibe trust of actual end-users. Finally, our experimental framework provides a principled approach for evaluating ``how congruent machine learning models are to the expectations of humans'', which can be defined in terms of ethical considerations.

\section{Conclusions and Future Work}
\label{sec:conc_fw}

In this paper, we focus on juxtaposing human understanding in an image recognition task with that of machines in two central scenarios of human decision making  -- \emph{ selection of discriminative segments in an image} and \emph{object recognition}. We conducted a large-scale crowd sourcing study entailing 7,000 HITs with an aim to further the understanding of dissonance between humans and machines in the image classification task. To this end, we proposed novel metrics to measure the dissonance between humans and 3 state-of-the-art neural network models (\inception{}, \resnet{}, \vgg{}). 
Our findings suggest that human perception of feature importance (i.e., the selection of discriminative segments in Task-1) does not consistently result in better human image recognition (in Task-2) in comparison to that by the neural network models considered in this work. We found that the models that are close to human understanding also generalise better.  %One of the key findings
Our experimental evidence shows that humans are not always able to effectively exploit the use of context towards determining good features (i.e., discriminative segments in images).  

		We also found that image difficulty is directly correlated with the effort in recognising objects irrespective of human or machine selected features. Finally, we release our entire dataset consisting of the two-stage crowd sourced tasks, complete with annotations from crowd workers for evaluation of image classification models. Our experiments in this paper shed light on the value that such a dataset and task design can bring to the CSCW and HCI community in furthering the understanding of human-machine dissonance. For example we unearth the fact that over-parametrized models like \vgg{} tend to be more robust even if they are not the best performing models in case of \textit{easy} images. We resonate that building more human-like machines can result in their seamless integration into our everyday lives, through interactions including collaboration and cooperation.  

		In our future work we will delve into investigating the dissonance between humans and machines (i) when they both make the same error (\textit{`are machines wrong for the right reasons?'}) which is key in critical domains like health and defence and (ii) in other tasks %designs for other machine learning tasks
		such as visual question answering, machine translation, document retrieval etc. We also aim to investigate effects of a closed domain assumption for image recognition and other classification tasks where the set of classes/labels are known to the assessor. 

\subsection*{Acknowledgements} We thank all the anonymous crowd workers who participated in our experiments. This research has been supported in part by the Amazon Research Awards, and the Erasmus+ project DISKOW (grant no. 60171990).

% \input{appendix}
%\newpage
\bibliographystyle{ACM-Reference-Format}
\bibliography{references}


\begin{thebibliography}{73}

%%% the Web address.

\ifx \showCODEN    \undefined \def \showCODEN     #1{\unskip}     \fi
\ifx \showDOI      \undefined \def \showDOI       #1{#1}\fi
\ifx \showISBNx    \undefined \def \showISBNx     #1{\unskip}     \fi
\ifx \showISBNxiii \undefined \def \showISBNxiii  #1{\unskip}     \fi
\ifx \showISSN     \undefined \def \showISSN      #1{\unskip}     \fi
\ifx \showLCCN     \undefined \def \showLCCN      #1{\unskip}     \fi
\ifx \shownote     \undefined \def \shownote      #1{#1}          \fi
\ifx \showarticletitle \undefined \def \showarticletitle #1{#1}   \fi
\ifx \showURL      \undefined \def \showURL       {\relax}        \fi
% invisible to TeX
\providecommand\bibfield[2]{#2}
\providecommand\bibinfo[2]{#2}
\providecommand\natexlab[1]{#1}
\providecommand\showeprint[2][]{arXiv:#2}

\bibitem[\protect\citeauthoryear{Abdul, Vermeulen, Wang, Lim, and
  Kankanhalli}{Abdul et~al\mbox{.}}{2018}]%
        {abdul2018trends}
\bibfield{author}{\bibinfo{person}{Ashraf Abdul}, \bibinfo{person}{Jo
  Vermeulen}, \bibinfo{person}{Danding Wang}, \bibinfo{person}{Brian~Y Lim},
  {and} \bibinfo{person}{Mohan Kankanhalli}.} \bibinfo{year}{2018}\natexlab{}.
\newblock \showarticletitle{Trends and trajectories for explainable,
  accountable and intelligible systems: An hci research agenda}. In
  \bibinfo{booktitle}{\emph{Proceedings of the 2018 CHI Conference on Human
  Factors in Computing Systems}}. ACM, \bibinfo{pages}{582}.
\newblock


\bibitem[\protect\citeauthoryear{Afraz, Yamins, and DiCarlo}{Afraz
  et~al\mbox{.}}{2014}]%
        {afraz2014neural}
\bibfield{author}{\bibinfo{person}{Arash Afraz}, \bibinfo{person}{Daniel~LK
  Yamins}, {and} \bibinfo{person}{James~J DiCarlo}.}
  \bibinfo{year}{2014}\natexlab{}.
\newblock \showarticletitle{Neural mechanisms underlying visual object
  recognition}. In \bibinfo{booktitle}{\emph{Cold Spring Harbor symposia on
  quantitative biology}}, Vol.~\bibinfo{volume}{79}. Cold Spring Harbor
  Laboratory Press, \bibinfo{pages}{99--107}.
\newblock


\bibitem[\protect\citeauthoryear{Anand, Bizer, Erlei, Gadiraju, Heinze, Meub,
  Nejdl, and Steinroetter}{Anand et~al\mbox{.}}{2018}]%
        {anand2018effects}
\bibfield{author}{\bibinfo{person}{Avishek Anand}, \bibinfo{person}{Kilian
  Bizer}, \bibinfo{person}{Alexander Erlei}, \bibinfo{person}{Ujwal Gadiraju},
  \bibinfo{person}{Christian Heinze}, \bibinfo{person}{Lukas Meub},
  \bibinfo{person}{Wolfgang Nejdl}, {and} \bibinfo{person}{Bjoern
  Steinroetter}.} \bibinfo{year}{2018}\natexlab{}.
\newblock \showarticletitle{Effects of Algorithmic Decision-Making and
  Interpretability on Human Behavior: Experiments using Crowdsourcing}. In
  \bibinfo{booktitle}{\emph{Proceedings of the {HCOMP} 2018 Works in Progress
  and Demonstration Papers Track of the sixth {AAAI} Conference on Human
  Computation and Crowdsourcing {(HCOMP} 2018), Zurich, Switzerland, July 5-8,
  2018.}}
\newblock


\bibitem[\protect\citeauthoryear{Auckland, Cave, and Donnelly}{Auckland
  et~al\mbox{.}}{2007}]%
        {auckland2007nontarget}
\bibfield{author}{\bibinfo{person}{Mark~E Auckland}, \bibinfo{person}{Kyle~R
  Cave}, {and} \bibinfo{person}{Nick Donnelly}.}
  \bibinfo{year}{2007}\natexlab{}.
\newblock \showarticletitle{Nontarget objects can influence perceptual
  processes during object recognition}.
\newblock \bibinfo{journal}{\emph{Psychonomic bulletin \& review}}
  \bibinfo{volume}{14}, \bibinfo{number}{2} (\bibinfo{year}{2007}),
  \bibinfo{pages}{332--337}.
\newblock


\bibitem[\protect\citeauthoryear{Berkovsky, Taib, and Conway}{Berkovsky
  et~al\mbox{.}}{2017}]%
        {berkovsky2017recommend}
\bibfield{author}{\bibinfo{person}{Shlomo Berkovsky}, \bibinfo{person}{Ronnie
  Taib}, {and} \bibinfo{person}{Dan Conway}.} \bibinfo{year}{2017}\natexlab{}.
\newblock \showarticletitle{How to recommend?: User trust factors in movie
  recommender systems}. In \bibinfo{booktitle}{\emph{Proceedings of the 22nd
  International Conference on Intelligent User Interfaces}}. ACM,
  \bibinfo{pages}{287--300}.
\newblock


\bibitem[\protect\citeauthoryear{Biederman}{Biederman}{1985}]%
        {biederman1985human}
\bibfield{author}{\bibinfo{person}{Irving Biederman}.}
  \bibinfo{year}{1985}\natexlab{}.
\newblock \showarticletitle{Human image understanding: Recent research and a
  theory}.
\newblock \bibinfo{journal}{\emph{Computer vision, graphics, and image
  processing}} \bibinfo{volume}{32}, \bibinfo{number}{1}
  (\bibinfo{year}{1985}), \bibinfo{pages}{29--73}.
\newblock


\bibitem[\protect\citeauthoryear{Binns, Van~Kleek, Veale, Lyngs, Zhao, and
  Shadbolt}{Binns et~al\mbox{.}}{2018}]%
        {Binns:2018:RHP:3173574.3173951}
\bibfield{author}{\bibinfo{person}{Reuben Binns}, \bibinfo{person}{Max
  Van~Kleek}, \bibinfo{person}{Michael Veale}, \bibinfo{person}{Ulrik Lyngs},
  \bibinfo{person}{Jun Zhao}, {and} \bibinfo{person}{Nigel Shadbolt}.}
  \bibinfo{year}{2018}\natexlab{}.
\newblock \showarticletitle{'It's Reducing a Human Being to a Percentage':
  Perceptions of Justice in Algorithmic Decisions}. In
  \bibinfo{booktitle}{\emph{Proceedings of the 2018 CHI Conference on Human
  Factors in Computing Systems}} \emph{(\bibinfo{series}{CHI '18})}.
  \bibinfo{publisher}{ACM}, \bibinfo{address}{New York, NY, USA}, Article
  \bibinfo{articleno}{377}, \bibinfo{numpages}{14}~pages.
\newblock
\showISBNx{978-1-4503-5620-6}
\urldef\tempurl%
\url{https://doi.org/10.1145/3173574.3173951}
\showDOI{\tempurl}


\bibitem[\protect\citeauthoryear{Caruana, Lou, Gehrke, Koch, Sturm, and
  Elhadad}{Caruana et~al\mbox{.}}{2015}]%
        {caruana2015intelligibletrees}
\bibfield{author}{\bibinfo{person}{Rich Caruana}, \bibinfo{person}{Yin Lou},
  \bibinfo{person}{Johannes Gehrke}, \bibinfo{person}{Paul Koch},
  \bibinfo{person}{Marc Sturm}, {and} \bibinfo{person}{Noemie Elhadad}.}
  \bibinfo{year}{2015}\natexlab{}.
\newblock \showarticletitle{Intelligible models for healthcare: Predicting
  pneumonia risk and hospital 30-day readmission}. In
  \bibinfo{booktitle}{\emph{Proceedings of the 21th ACM SIGKDD International
  Conference on Knowledge Discovery and Data Mining}}. ACM,
  \bibinfo{pages}{1721--1730}.
\newblock


\bibitem[\protect\citeauthoryear{Deng, Dong, Socher, Li, Li, and Fei-Fei}{Deng
  et~al\mbox{.}}{2009}]%
        {deng2009imagenet}
\bibfield{author}{\bibinfo{person}{Jia Deng}, \bibinfo{person}{Wei Dong},
  \bibinfo{person}{Richard Socher}, \bibinfo{person}{Li-Jia Li},
  \bibinfo{person}{Kai Li}, {and} \bibinfo{person}{Li Fei-Fei}.}
  \bibinfo{year}{2009}\natexlab{}.
\newblock \showarticletitle{Imagenet: A large-scale hierarchical image
  database}. In \bibinfo{booktitle}{\emph{Computer Vision and Pattern
  Recognition, 2009. CVPR 2009. IEEE Conference on}}. Ieee,
  \bibinfo{pages}{248--255}.
\newblock


\bibitem[\protect\citeauthoryear{Doshi-Velez and Kim}{Doshi-Velez and
  Kim}{2017}]%
        {doshi2017towards}
\bibfield{author}{\bibinfo{person}{Finale Doshi-Velez} {and}
  \bibinfo{person}{Been Kim}.} \bibinfo{year}{2017}\natexlab{}.
\newblock \showarticletitle{Towards a rigorous science of interpretable machine
  learning}.
\newblock  (\bibinfo{year}{2017}).
\newblock


\bibitem[\protect\citeauthoryear{Doumas, Puebla, and Martin}{Doumas
  et~al\mbox{.}}{2018}]%
        {doumas2018human}
\bibfield{author}{\bibinfo{person}{Leonidas~AA Doumas},
  \bibinfo{person}{Guillermo Puebla}, {and} \bibinfo{person}{Andrea~E Martin}.}
  \bibinfo{year}{2018}\natexlab{}.
\newblock \showarticletitle{Human-like generalization in a machine through
  predicate learning}.
\newblock \bibinfo{journal}{\emph{arXiv preprint arXiv:1806.01709}}
  (\bibinfo{year}{2018}).
\newblock


\bibitem[\protect\citeauthoryear{Eysenck and Keane}{Eysenck and Keane}{2013}]%
        {eysenck2013cognitive}
\bibfield{author}{\bibinfo{person}{Michael~W Eysenck} {and}
  \bibinfo{person}{Mark~T Keane}.} \bibinfo{year}{2013}\natexlab{}.
\newblock \bibinfo{booktitle}{\emph{Cognitive psychology: A student's
  handbook}}.
\newblock \bibinfo{publisher}{Psychology press}.
\newblock


\bibitem[\protect\citeauthoryear{Friedrich and Zanker}{Friedrich and
  Zanker}{2011}]%
        {friedrich2011taxonomy}
\bibfield{author}{\bibinfo{person}{Gerhard Friedrich} {and}
  \bibinfo{person}{Markus Zanker}.} \bibinfo{year}{2011}\natexlab{}.
\newblock \showarticletitle{A taxonomy for generating explanations in
  recommender systems}.
\newblock \bibinfo{journal}{\emph{AI Magazine}} \bibinfo{volume}{32},
  \bibinfo{number}{3} (\bibinfo{year}{2011}), \bibinfo{pages}{90--98}.
\newblock


\bibitem[\protect\citeauthoryear{Gadiraju, Fetahu, and Kawase}{Gadiraju
  et~al\mbox{.}}{2015a}]%
        {gadiraju2015training}
\bibfield{author}{\bibinfo{person}{Ujwal Gadiraju}, \bibinfo{person}{Besnik
  Fetahu}, {and} \bibinfo{person}{Ricardo Kawase}.}
  \bibinfo{year}{2015}\natexlab{a}.
\newblock \showarticletitle{Training workers for improving performance in
  crowdsourcing microtasks}.
\newblock In \bibinfo{booktitle}{\emph{Design for Teaching and Learning in a
  Networked World}}. \bibinfo{publisher}{Springer}, \bibinfo{pages}{100--114}.
\newblock


\bibitem[\protect\citeauthoryear{Gadiraju, Kawase, Dietze, and
  Demartini}{Gadiraju et~al\mbox{.}}{2015b}]%
        {gadiraju2015understanding}
\bibfield{author}{\bibinfo{person}{Ujwal Gadiraju}, \bibinfo{person}{Ricardo
  Kawase}, \bibinfo{person}{Stefan Dietze}, {and} \bibinfo{person}{Gianluca
  Demartini}.} \bibinfo{year}{2015}\natexlab{b}.
\newblock \showarticletitle{Understanding malicious behavior in crowdsourcing
  platforms: The case of online surveys}. In
  \bibinfo{booktitle}{\emph{Proceedings of the 33rd Annual ACM Conference on
  Human Factors in Computing Systems}}. ACM, \bibinfo{pages}{1631--1640}.
\newblock


\bibitem[\protect\citeauthoryear{Gadiraju, Yang, and Bozzon}{Gadiraju
  et~al\mbox{.}}{2017}]%
        {gadiraju2017clarity}
\bibfield{author}{\bibinfo{person}{Ujwal Gadiraju}, \bibinfo{person}{Jie Yang},
  {and} \bibinfo{person}{Alessandro Bozzon}.} \bibinfo{year}{2017}\natexlab{}.
\newblock \showarticletitle{Clarity is a worthwhile quality: On the role of
  task clarity in microtask crowdsourcing}. In
  \bibinfo{booktitle}{\emph{Proceedings of the 28th ACM Conference on Hypertext
  and Social Media}}. ACM, \bibinfo{pages}{5--14}.
\newblock


\bibitem[\protect\citeauthoryear{Geirhos, Temme, Rauber, Sch{\"u}tt, Bethge,
  and Wichmann}{Geirhos et~al\mbox{.}}{2018}]%
        {geirhos2018generalisation}
\bibfield{author}{\bibinfo{person}{Robert Geirhos}, \bibinfo{person}{Carlos~RM
  Temme}, \bibinfo{person}{Jonas Rauber}, \bibinfo{person}{Heiko~H Sch{\"u}tt},
  \bibinfo{person}{Matthias Bethge}, {and} \bibinfo{person}{Felix~A Wichmann}.}
  \bibinfo{year}{2018}\natexlab{}.
\newblock \showarticletitle{Generalisation in humans and deep neural networks}.
  In \bibinfo{booktitle}{\emph{Advances in Neural Information Processing
  Systems}}. \bibinfo{pages}{7549--7561}.
\newblock


\bibitem[\protect\citeauthoryear{Giboney, Brown, Lowry, and
  Nunamaker~Jr}{Giboney et~al\mbox{.}}{2015}]%
        {giboney2015user}
\bibfield{author}{\bibinfo{person}{Justin~Scott Giboney},
  \bibinfo{person}{Susan~A Brown}, \bibinfo{person}{Paul~Benjamin Lowry}, {and}
  \bibinfo{person}{Jay~F Nunamaker~Jr}.} \bibinfo{year}{2015}\natexlab{}.
\newblock \showarticletitle{User acceptance of knowledge-based system
  recommendations: Explanations, arguments, and fit}.
\newblock \bibinfo{journal}{\emph{Decision Support Systems}}
  \bibinfo{volume}{72} (\bibinfo{year}{2015}), \bibinfo{pages}{1--10}.
\newblock


\bibitem[\protect\citeauthoryear{Gregor and Benbasat}{Gregor and
  Benbasat}{1999}]%
        {gregor1999explanations}
\bibfield{author}{\bibinfo{person}{Shirley Gregor} {and} \bibinfo{person}{Izak
  Benbasat}.} \bibinfo{year}{1999}\natexlab{}.
\newblock \showarticletitle{Explanations from intelligent systems: Theoretical
  foundations and implications for practice}.
\newblock \bibinfo{journal}{\emph{MIS quarterly}} (\bibinfo{year}{1999}),
  \bibinfo{pages}{497--530}.
\newblock


\bibitem[\protect\citeauthoryear{Hann{\'a}k, Wagner, Garcia, Mislove,
  Strohmaier, and Wilson}{Hann{\'a}k et~al\mbox{.}}{2017}]%
        {hannak2017bias}
\bibfield{author}{\bibinfo{person}{Anik{\'o} Hann{\'a}k},
  \bibinfo{person}{Claudia Wagner}, \bibinfo{person}{David Garcia},
  \bibinfo{person}{Alan Mislove}, \bibinfo{person}{Markus Strohmaier}, {and}
  \bibinfo{person}{Christo Wilson}.} \bibinfo{year}{2017}\natexlab{}.
\newblock \showarticletitle{Bias in online freelance marketplaces: Evidence
  from taskrabbit and fiverr}. In \bibinfo{booktitle}{\emph{Proceedings of the
  2017 ACM Conference on Computer Supported Cooperative Work and Social
  Computing}}. ACM, \bibinfo{pages}{1914--1933}.
\newblock


\bibitem[\protect\citeauthoryear{Initiative et~al\mbox{.}}{Initiative
  et~al\mbox{.}}{2016}]%
        {ieee2016ethically}
\bibfield{author}{\bibinfo{person}{IEEE~Global Initiative} {et~al\mbox{.}}}
  \bibinfo{year}{2016}\natexlab{}.
\newblock \showarticletitle{Ethically Aligned Design}.
\newblock \bibinfo{journal}{\emph{IEEE Standards v1}} (\bibinfo{year}{2016}).
\newblock


\bibitem[\protect\citeauthoryear{J{\"a}rvelin and
  Kek{\"a}l{\"a}inen}{J{\"a}rvelin and Kek{\"a}l{\"a}inen}{2002}]%
        {jarvelin2002cumulated}
\bibfield{author}{\bibinfo{person}{Kalervo J{\"a}rvelin} {and}
  \bibinfo{person}{Jaana Kek{\"a}l{\"a}inen}.} \bibinfo{year}{2002}\natexlab{}.
\newblock \showarticletitle{Cumulated gain-based evaluation of IR techniques}.
\newblock \bibinfo{journal}{\emph{ACM Transactions on Information Systems
  (TOIS)}} \bibinfo{volume}{20}, \bibinfo{number}{4} (\bibinfo{year}{2002}),
  \bibinfo{pages}{422--446}.
\newblock


\bibitem[\protect\citeauthoryear{Josephy, Lease, Paritosh, Krause, Georgescu,
  Tjalve, and Braga}{Josephy et~al\mbox{.}}{2014}]%
        {crowdscale2013}
\bibfield{author}{\bibinfo{person}{Tatiana Josephy}, \bibinfo{person}{Matt
  Lease}, \bibinfo{person}{Praveen Paritosh}, \bibinfo{person}{Markus Krause},
  \bibinfo{person}{Mihai Georgescu}, \bibinfo{person}{Michael Tjalve}, {and}
  \bibinfo{person}{Daniela Braga}.} \bibinfo{year}{2014}\natexlab{}.
\newblock \showarticletitle{CrowdScale 2013: Crowdsourcing at Scale Workshop
  Report}.
\newblock \bibinfo{journal}{\emph{AI Magazine}} \bibinfo{volume}{35},
  \bibinfo{number}{2} (\bibinfo{year}{2014}), \bibinfo{pages}{75--78}.
\newblock


\bibitem[\protect\citeauthoryear{Kahneman}{Kahneman}{2003}]%
        {kahneman2003perspective}
\bibfield{author}{\bibinfo{person}{Daniel Kahneman}.}
  \bibinfo{year}{2003}\natexlab{}.
\newblock \showarticletitle{A perspective on judgment and choice: mapping
  bounded rationality.}
\newblock \bibinfo{journal}{\emph{American psychologist}} \bibinfo{volume}{58},
  \bibinfo{number}{9} (\bibinfo{year}{2003}), \bibinfo{pages}{697}.
\newblock


\bibitem[\protect\citeauthoryear{Kahneman, Rosenfield, Gandhi, and
  Blaser}{Kahneman et~al\mbox{.}}{2016}]%
        {kahneman2016noise}
\bibfield{author}{\bibinfo{person}{Daniel Kahneman}, \bibinfo{person}{Andrew~M
  Rosenfield}, \bibinfo{person}{Linnea Gandhi}, {and} \bibinfo{person}{Tom
  Blaser}.} \bibinfo{year}{2016}\natexlab{}.
\newblock \showarticletitle{Noise: How to overcome the high, hidden cost of
  inconsistent decision making}.
\newblock \bibinfo{journal}{\emph{Harvard business review}}
  \bibinfo{volume}{94}, \bibinfo{number}{10} (\bibinfo{year}{2016}),
  \bibinfo{pages}{38--46}.
\newblock


\bibitem[\protect\citeauthoryear{Kim, Khanna, and Koyejo}{Kim
  et~al\mbox{.}}{2016}]%
        {kim2016examples}
\bibfield{author}{\bibinfo{person}{Been Kim}, \bibinfo{person}{Rajiv Khanna},
  {and} \bibinfo{person}{Oluwasanmi~O Koyejo}.}
  \bibinfo{year}{2016}\natexlab{}.
\newblock \showarticletitle{Examples are not enough, learn to criticize!
  criticism for interpretability}. In \bibinfo{booktitle}{\emph{Advances in
  Neural Information Processing Systems}}. \bibinfo{pages}{2280--2288}.
\newblock


\bibitem[\protect\citeauthoryear{Kleinberg, Lakkaraju, Leskovec, Ludwig, and
  Mullainathan}{Kleinberg et~al\mbox{.}}{2017}]%
        {kleinberg2017human}
\bibfield{author}{\bibinfo{person}{Jon Kleinberg}, \bibinfo{person}{Himabindu
  Lakkaraju}, \bibinfo{person}{Jure Leskovec}, \bibinfo{person}{Jens Ludwig},
  {and} \bibinfo{person}{Sendhil Mullainathan}.}
  \bibinfo{year}{2017}\natexlab{}.
\newblock \showarticletitle{Human decisions and machine predictions}.
\newblock \bibinfo{journal}{\emph{The quarterly journal of economics}}
  \bibinfo{volume}{133}, \bibinfo{number}{1} (\bibinfo{year}{2017}),
  \bibinfo{pages}{237--293}.
\newblock


\bibitem[\protect\citeauthoryear{Koh and Liang}{Koh and Liang}{2017}]%
        {influencefunctionskoh2017understanding}
\bibfield{author}{\bibinfo{person}{Pang~Wei Koh} {and} \bibinfo{person}{Percy
  Liang}.} \bibinfo{year}{2017}\natexlab{}.
\newblock \showarticletitle{Understanding black-box predictions via influence
  functions}.
\newblock \bibinfo{journal}{\emph{arXiv preprint arXiv:1703.04730}}
  (\bibinfo{year}{2017}).
\newblock


\bibitem[\protect\citeauthoryear{Krishna, Hata, Chen, Kravitz, Shamma, Fei-Fei,
  and Bernstein}{Krishna et~al\mbox{.}}{2016}]%
        {krishna2016embracing}
\bibfield{author}{\bibinfo{person}{Ranjay~A Krishna}, \bibinfo{person}{Kenji
  Hata}, \bibinfo{person}{Stephanie Chen}, \bibinfo{person}{Joshua Kravitz},
  \bibinfo{person}{David~A Shamma}, \bibinfo{person}{Li Fei-Fei}, {and}
  \bibinfo{person}{Michael~S Bernstein}.} \bibinfo{year}{2016}\natexlab{}.
\newblock \showarticletitle{Embracing error to enable rapid crowdsourcing}. In
  \bibinfo{booktitle}{\emph{Proceedings of the 2016 CHI conference on human
  factors in computing systems}}. ACM, \bibinfo{pages}{3167--3179}.
\newblock


\bibitem[\protect\citeauthoryear{Krizhevsky, Sutskever, and Hinton}{Krizhevsky
  et~al\mbox{.}}{2012}]%
        {krizhevsky2012imagenet}
\bibfield{author}{\bibinfo{person}{Alex Krizhevsky}, \bibinfo{person}{Ilya
  Sutskever}, {and} \bibinfo{person}{Geoffrey~E Hinton}.}
  \bibinfo{year}{2012}\natexlab{}.
\newblock \showarticletitle{Imagenet classification with deep convolutional
  neural networks}. In \bibinfo{booktitle}{\emph{Advances in neural information
  processing systems}}. \bibinfo{pages}{1097--1105}.
\newblock


\bibitem[\protect\citeauthoryear{Lake, Ullman, Tenenbaum, and Gershman}{Lake
  et~al\mbox{.}}{2017}]%
        {lake2017building}
\bibfield{author}{\bibinfo{person}{Brenden~M Lake}, \bibinfo{person}{Tomer~D
  Ullman}, \bibinfo{person}{Joshua~B Tenenbaum}, {and}
  \bibinfo{person}{Samuel~J Gershman}.} \bibinfo{year}{2017}\natexlab{}.
\newblock \showarticletitle{Building machines that learn and think like
  people}.
\newblock \bibinfo{journal}{\emph{Behavioral and Brain Sciences}}
  \bibinfo{volume}{40} (\bibinfo{year}{2017}).
\newblock


\bibitem[\protect\citeauthoryear{Lawson, Hiatt, and Trafton}{Lawson
  et~al\mbox{.}}{2014}]%
        {lawson2014leveraging}
\bibfield{author}{\bibinfo{person}{Wallace Lawson}, \bibinfo{person}{Laura
  Hiatt}, {and} \bibinfo{person}{J Trafton}.} \bibinfo{year}{2014}\natexlab{}.
\newblock \showarticletitle{Leveraging cognitive context for object
  recognition}. In \bibinfo{booktitle}{\emph{Proceedings of the IEEE Conference
  on Computer Vision and Pattern Recognition Workshops}}.
  \bibinfo{pages}{381--386}.
\newblock


\bibitem[\protect\citeauthoryear{Lee and Baykal}{Lee and Baykal}{2017}]%
        {lee2017algorithmic}
\bibfield{author}{\bibinfo{person}{Min~Kyung Lee} {and} \bibinfo{person}{Su
  Baykal}.} \bibinfo{year}{2017}\natexlab{}.
\newblock \showarticletitle{Algorithmic mediation in group decisions: Fairness
  perceptions of algorithmically mediated vs. discussion-based social
  division}. In \bibinfo{booktitle}{\emph{Proceedings of the 2017 ACM
  Conference on Computer Supported Cooperative Work and Social Computing}}.
  ACM, \bibinfo{pages}{1035--1048}.
\newblock


\bibitem[\protect\citeauthoryear{Lee, Kusbit, Metsky, and Dabbish}{Lee
  et~al\mbox{.}}{2015}]%
        {lee2015working}
\bibfield{author}{\bibinfo{person}{Min~Kyung Lee}, \bibinfo{person}{Daniel
  Kusbit}, \bibinfo{person}{Evan Metsky}, {and} \bibinfo{person}{Laura
  Dabbish}.} \bibinfo{year}{2015}\natexlab{}.
\newblock \showarticletitle{Working with machines: The impact of algorithmic
  and data-driven management on human workers}. In
  \bibinfo{booktitle}{\emph{Proceedings of the 33rd Annual ACM Conference on
  Human Factors in Computing Systems}}. ACM, \bibinfo{pages}{1603--1612}.
\newblock


\bibitem[\protect\citeauthoryear{Letham, Rudin, McCormick, Madigan,
  et~al\mbox{.}}{Letham et~al\mbox{.}}{2015}]%
        {rulesletham2015interpretable}
\bibfield{author}{\bibinfo{person}{Benjamin Letham}, \bibinfo{person}{Cynthia
  Rudin}, \bibinfo{person}{Tyler~H McCormick}, \bibinfo{person}{David Madigan},
  {et~al\mbox{.}}} \bibinfo{year}{2015}\natexlab{}.
\newblock \showarticletitle{Interpretable classifiers using rules and Bayesian
  analysis: Building a better stroke prediction model}.
\newblock \bibinfo{journal}{\emph{The Annals of Applied Statistics}}
  \bibinfo{volume}{9}, \bibinfo{number}{3} (\bibinfo{year}{2015}),
  \bibinfo{pages}{1350--1371}.
\newblock


\bibitem[\protect\citeauthoryear{Lipton}{Lipton}{2016}]%
        {lipton2016mythos}
\bibfield{author}{\bibinfo{person}{Zachary~C Lipton}.}
  \bibinfo{year}{2016}\natexlab{}.
\newblock \showarticletitle{The mythos of model interpretability}.
\newblock \bibinfo{journal}{\emph{ICML Workshop on Human Interpretability of
  Machine Learning}} (\bibinfo{year}{2016}).
\newblock


\bibitem[\protect\citeauthoryear{Lundberg and Lee}{Lundberg and Lee}{2017}]%
        {lundberg2017unified}
\bibfield{author}{\bibinfo{person}{Scott~M Lundberg} {and}
  \bibinfo{person}{Su-In Lee}.} \bibinfo{year}{2017}\natexlab{}.
\newblock \showarticletitle{A unified approach to interpreting model
  predictions}. In \bibinfo{booktitle}{\emph{Advances in Neural Information
  Processing Systems}}. \bibinfo{pages}{4765--4774}.
\newblock


\bibitem[\protect\citeauthoryear{Monge}{Monge}{1781}]%
        {monge1781memoire}
\bibfield{author}{\bibinfo{person}{Gaspard Monge}.}
  \bibinfo{year}{1781}\natexlab{}.
\newblock \showarticletitle{M{\'e}moire sur la th{\'e}orie des d{\'e}blais et
  des remblais}.
\newblock \bibinfo{journal}{\emph{Histoire de l'Acad{\'e}mie Royale des
  Sciences de Paris}} (\bibinfo{year}{1781}).
\newblock


\bibitem[\protect\citeauthoryear{Myers}{Myers}{2002}]%
        {myers2002powers}
\bibfield{author}{\bibinfo{person}{David~G Myers}.}
  \bibinfo{year}{2002}\natexlab{}.
\newblock \showarticletitle{The powers \& perils of intuition}.
\newblock \bibinfo{journal}{\emph{Psychology Today}} \bibinfo{volume}{35},
  \bibinfo{number}{6} (\bibinfo{year}{2002}), \bibinfo{pages}{42--52}.
\newblock


\bibitem[\protect\citeauthoryear{Oduor and Wiebe}{Oduor and Wiebe}{2008}]%
        {oduor2008effects}
\bibfield{author}{\bibinfo{person}{Kenya~Freeman Oduor} {and}
  \bibinfo{person}{Eric~N Wiebe}.} \bibinfo{year}{2008}\natexlab{}.
\newblock \showarticletitle{The effects of automated decision algorithm
  modality and transparency on reported trust and task performance}. In
  \bibinfo{booktitle}{\emph{Proceedings of the Human Factors and Ergonomics
  Society Annual Meeting}}, Vol.~\bibinfo{volume}{52}. SAGE Publications Sage
  CA: Los Angeles, CA, \bibinfo{pages}{302--306}.
\newblock


\bibitem[\protect\citeauthoryear{Oleson, Sorokin, Laughlin, Hester, Le, and
  Biewald}{Oleson et~al\mbox{.}}{2011}]%
        {oleson2011programmatic}
\bibfield{author}{\bibinfo{person}{David Oleson}, \bibinfo{person}{Alexander
  Sorokin}, \bibinfo{person}{Greg~P Laughlin}, \bibinfo{person}{Vaughn Hester},
  \bibinfo{person}{John Le}, {and} \bibinfo{person}{Lukas Biewald}.}
  \bibinfo{year}{2011}\natexlab{}.
\newblock \showarticletitle{Programmatic Gold: Targeted and Scalable Quality
  Assurance in Crowdsourcing.}
\newblock \bibinfo{journal}{\emph{Human computation}} \bibinfo{volume}{11},
  \bibinfo{number}{11} (\bibinfo{year}{2011}).
\newblock


\bibitem[\protect\citeauthoryear{O’Neill}{O’Neill}{2016}]%
        {o2016weapons}
\bibfield{author}{\bibinfo{person}{Cathy O’Neill}.}
  \bibinfo{year}{2016}\natexlab{}.
\newblock \showarticletitle{Weapons of math destruction: How big data increases
  inequality and threatens democracy}.
\newblock \bibinfo{journal}{\emph{Nueva York, NY: Crown Publishing Group}}
  (\bibinfo{year}{2016}).
\newblock


\bibitem[\protect\citeauthoryear{Papadimitriou, Symeonidis, and
  Manolopoulos}{Papadimitriou et~al\mbox{.}}{2012}]%
        {papadimitriou2012generalized}
\bibfield{author}{\bibinfo{person}{Alexis Papadimitriou},
  \bibinfo{person}{Panagiotis Symeonidis}, {and} \bibinfo{person}{Yannis
  Manolopoulos}.} \bibinfo{year}{2012}\natexlab{}.
\newblock \showarticletitle{A generalized taxonomy of explanations styles for
  traditional and social recommender systems}.
\newblock \bibinfo{journal}{\emph{Data Mining and Knowledge Discovery}}
  \bibinfo{volume}{24}, \bibinfo{number}{3} (\bibinfo{year}{2012}),
  \bibinfo{pages}{555--583}.
\newblock


\bibitem[\protect\citeauthoryear{Plackett}{Plackett}{1975}]%
        {plackett1975analysis}
\bibfield{author}{\bibinfo{person}{Robin~L Plackett}.}
  \bibinfo{year}{1975}\natexlab{}.
\newblock \showarticletitle{The analysis of permutations}.
\newblock \bibinfo{journal}{\emph{Applied Statistics}} (\bibinfo{year}{1975}),
  \bibinfo{pages}{193--202}.
\newblock


\bibitem[\protect\citeauthoryear{Porcheron, Fischer, Reeves, and
  Sharples}{Porcheron et~al\mbox{.}}{2018}]%
        {porcheron2018voice}
\bibfield{author}{\bibinfo{person}{Martin Porcheron}, \bibinfo{person}{Joel~E
  Fischer}, \bibinfo{person}{Stuart Reeves}, {and} \bibinfo{person}{Sarah
  Sharples}.} \bibinfo{year}{2018}\natexlab{}.
\newblock \showarticletitle{Voice interfaces in everyday life}. In
  \bibinfo{booktitle}{\emph{proceedings of the 2018 CHI conference on human
  factors in computing systems}}. ACM, \bibinfo{pages}{640}.
\newblock


\bibitem[\protect\citeauthoryear{Rader, Cotter, and Cho}{Rader
  et~al\mbox{.}}{2018}]%
        {Rader:2018:EMS:3173574.3173677}
\bibfield{author}{\bibinfo{person}{Emilee Rader}, \bibinfo{person}{Kelley
  Cotter}, {and} \bibinfo{person}{Janghee Cho}.}
  \bibinfo{year}{2018}\natexlab{}.
\newblock \showarticletitle{Explanations As Mechanisms for Supporting
  Algorithmic Transparency}. In \bibinfo{booktitle}{\emph{Proceedings of the
  2018 CHI Conference on Human Factors in Computing Systems}}
  \emph{(\bibinfo{series}{CHI '18})}. \bibinfo{publisher}{ACM},
  \bibinfo{address}{New York, NY, USA}, Article \bibinfo{articleno}{103},
  \bibinfo{numpages}{13}~pages.
\newblock
\showISBNx{978-1-4503-5620-6}
\urldef\tempurl%
\url{https://doi.org/10.1145/3173574.3173677}
\showDOI{\tempurl}


\bibitem[\protect\citeauthoryear{Rahwan, Cebrian, Obradovich, Bongard,
  Bonnefon, Breazeal, Crandall, Christakis, Couzin, Jackson,
  et~al\mbox{.}}{Rahwan et~al\mbox{.}}{2019}]%
        {rahwan2019machine}
\bibfield{author}{\bibinfo{person}{Iyad Rahwan}, \bibinfo{person}{Manuel
  Cebrian}, \bibinfo{person}{Nick Obradovich}, \bibinfo{person}{Josh Bongard},
  \bibinfo{person}{Jean-Fran{\c{c}}ois Bonnefon}, \bibinfo{person}{Cynthia
  Breazeal}, \bibinfo{person}{Jacob~W Crandall}, \bibinfo{person}{Nicholas~A
  Christakis}, \bibinfo{person}{Iain~D Couzin}, \bibinfo{person}{Matthew~O
  Jackson}, {et~al\mbox{.}}} \bibinfo{year}{2019}\natexlab{}.
\newblock \showarticletitle{Machine behaviour}.
\newblock \bibinfo{journal}{\emph{Nature}} \bibinfo{volume}{568},
  \bibinfo{number}{7753} (\bibinfo{year}{2019}), \bibinfo{pages}{477}.
\newblock


\bibitem[\protect\citeauthoryear{Rajalingham, Issa, Bashivan, Kar, Schmidt, and
  DiCarlo}{Rajalingham et~al\mbox{.}}{2018}]%
        {rajalingham2018large}
\bibfield{author}{\bibinfo{person}{Rishi Rajalingham}, \bibinfo{person}{Elias~B
  Issa}, \bibinfo{person}{Pouya Bashivan}, \bibinfo{person}{Kohitij Kar},
  \bibinfo{person}{Kailyn Schmidt}, {and} \bibinfo{person}{James~J DiCarlo}.}
  \bibinfo{year}{2018}\natexlab{}.
\newblock \showarticletitle{Large-scale, high-resolution comparison of the core
  visual object recognition behavior of humans, monkeys, and state-of-the-art
  deep artificial neural networks}.
\newblock \bibinfo{journal}{\emph{Journal of Neuroscience}}
  \bibinfo{volume}{38}, \bibinfo{number}{33} (\bibinfo{year}{2018}),
  \bibinfo{pages}{7255--7269}.
\newblock


\bibitem[\protect\citeauthoryear{Ribeiro, Singh, and Guestrin}{Ribeiro
  et~al\mbox{.}}{2016a}]%
        {ribeiro2016model}
\bibfield{author}{\bibinfo{person}{Marco~Tulio Ribeiro},
  \bibinfo{person}{Sameer Singh}, {and} \bibinfo{person}{Carlos Guestrin}.}
  \bibinfo{year}{2016}\natexlab{a}.
\newblock \showarticletitle{Model-agnostic interpretability of machine
  learning}.
\newblock \bibinfo{journal}{\emph{arXiv preprint arXiv:1606.05386}}
  (\bibinfo{year}{2016}).
\newblock


\bibitem[\protect\citeauthoryear{Ribeiro, Singh, and Guestrin}{Ribeiro
  et~al\mbox{.}}{2016b}]%
        {ribeiro2016should}
\bibfield{author}{\bibinfo{person}{Marco~Tulio Ribeiro},
  \bibinfo{person}{Sameer Singh}, {and} \bibinfo{person}{Carlos Guestrin}.}
  \bibinfo{year}{2016}\natexlab{b}.
\newblock \showarticletitle{Why Should I Trust You?: Explaining the Predictions
  of Any Classifier}. In \bibinfo{booktitle}{\emph{Proceedings of the 22nd ACM
  SIGKDD International Conference on Knowledge Discovery and Data Mining}}.
  ACM, \bibinfo{pages}{1135--1144}.
\newblock


\bibitem[\protect\citeauthoryear{Ross, Hughes, and Doshi-Velez}{Ross
  et~al\mbox{.}}{2017}]%
        {ijcai2017-371}
\bibfield{author}{\bibinfo{person}{Andrew~Slavin Ross},
  \bibinfo{person}{Michael~C. Hughes}, {and} \bibinfo{person}{Finale
  Doshi-Velez}.} \bibinfo{year}{2017}\natexlab{}.
\newblock \showarticletitle{Right for the Right Reasons: Training
  Differentiable Models by Constraining their Explanations}. In
  \bibinfo{booktitle}{\emph{Proceedings of the Twenty-Sixth International Joint
  Conference on Artificial Intelligence, {IJCAI-17}}}.
  \bibinfo{pages}{2662--2670}.
\newblock
\urldef\tempurl%
\url{https://doi.org/10.24963/ijcai.2017/371}
\showDOI{\tempurl}


\bibitem[\protect\citeauthoryear{Rubner, Tomasi, and Guibas}{Rubner
  et~al\mbox{.}}{1998}]%
        {rubner1998metric}
\bibfield{author}{\bibinfo{person}{Yossi Rubner}, \bibinfo{person}{Carlo
  Tomasi}, {and} \bibinfo{person}{Leonidas~J Guibas}.}
  \bibinfo{year}{1998}\natexlab{}.
\newblock \showarticletitle{A metric for distributions with applications to
  image databases}. In \bibinfo{booktitle}{\emph{Computer Vision, 1998. Sixth
  International Conference on}}. IEEE, \bibinfo{pages}{59--66}.
\newblock


\bibitem[\protect\citeauthoryear{Schrimpf, Kubilius, Hong, Majaj, Rajalingham,
  Issa, Kar, Bashivan, Prescott-Roy, Schmidt, et~al\mbox{.}}{Schrimpf
  et~al\mbox{.}}{2018}]%
        {schrimpf2018brain}
\bibfield{author}{\bibinfo{person}{Martin Schrimpf}, \bibinfo{person}{Jonas
  Kubilius}, \bibinfo{person}{Ha Hong}, \bibinfo{person}{Najib~J Majaj},
  \bibinfo{person}{Rishi Rajalingham}, \bibinfo{person}{Elias~B Issa},
  \bibinfo{person}{Kohitij Kar}, \bibinfo{person}{Pouya Bashivan},
  \bibinfo{person}{Jonathan Prescott-Roy}, \bibinfo{person}{Kailyn Schmidt},
  {et~al\mbox{.}}} \bibinfo{year}{2018}\natexlab{}.
\newblock \showarticletitle{Brain-Score: Which Artificial Neural Network for
  Object Recognition is most Brain-Like?}
\newblock \bibinfo{journal}{\emph{BioRxiv}} (\bibinfo{year}{2018}),
  \bibinfo{pages}{407007}.
\newblock


\bibitem[\protect\citeauthoryear{Shieh}{Shieh}{1998}]%
        {shieh1998weighted}
\bibfield{author}{\bibinfo{person}{Grace~S Shieh}.}
  \bibinfo{year}{1998}\natexlab{}.
\newblock \showarticletitle{A weighted Kendall's tau statistic}.
\newblock \bibinfo{journal}{\emph{Statistics \& probability letters}}
  \bibinfo{volume}{39}, \bibinfo{number}{1} (\bibinfo{year}{1998}),
  \bibinfo{pages}{17--24}.
\newblock


\bibitem[\protect\citeauthoryear{Shirado and Christakis}{Shirado and
  Christakis}{2017}]%
        {shirado2017locally}
\bibfield{author}{\bibinfo{person}{Hirokazu Shirado} {and}
  \bibinfo{person}{Nicholas~A Christakis}.} \bibinfo{year}{2017}\natexlab{}.
\newblock \showarticletitle{Locally noisy autonomous agents improve global
  human coordination in network experiments}.
\newblock \bibinfo{journal}{\emph{Nature}} \bibinfo{volume}{545},
  \bibinfo{number}{7654} (\bibinfo{year}{2017}), \bibinfo{pages}{370}.
\newblock


\bibitem[\protect\citeauthoryear{Silver, Huang, Maddison, Guez, Sifre, Van
  Den~Driessche, Schrittwieser, Antonoglou, Panneershelvam, Lanctot,
  et~al\mbox{.}}{Silver et~al\mbox{.}}{2016}]%
        {silver2016mastering}
\bibfield{author}{\bibinfo{person}{David Silver}, \bibinfo{person}{Aja Huang},
  \bibinfo{person}{Chris~J Maddison}, \bibinfo{person}{Arthur Guez},
  \bibinfo{person}{Laurent Sifre}, \bibinfo{person}{George Van Den~Driessche},
  \bibinfo{person}{Julian Schrittwieser}, \bibinfo{person}{Ioannis Antonoglou},
  \bibinfo{person}{Veda Panneershelvam}, \bibinfo{person}{Marc Lanctot},
  {et~al\mbox{.}}} \bibinfo{year}{2016}\natexlab{}.
\newblock \showarticletitle{Mastering the game of Go with deep neural networks
  and tree search}.
\newblock \bibinfo{journal}{\emph{nature}} \bibinfo{volume}{529},
  \bibinfo{number}{7587} (\bibinfo{year}{2016}), \bibinfo{pages}{484}.
\newblock


\bibitem[\protect\citeauthoryear{Silver, Schrittwieser, Simonyan, Antonoglou,
  Huang, Guez, Hubert, Baker, Lai, Bolton, et~al\mbox{.}}{Silver
  et~al\mbox{.}}{2017}]%
        {silver2017mastering}
\bibfield{author}{\bibinfo{person}{David Silver}, \bibinfo{person}{Julian
  Schrittwieser}, \bibinfo{person}{Karen Simonyan}, \bibinfo{person}{Ioannis
  Antonoglou}, \bibinfo{person}{Aja Huang}, \bibinfo{person}{Arthur Guez},
  \bibinfo{person}{Thomas Hubert}, \bibinfo{person}{Lucas Baker},
  \bibinfo{person}{Matthew Lai}, \bibinfo{person}{Adrian Bolton},
  {et~al\mbox{.}}} \bibinfo{year}{2017}\natexlab{}.
\newblock \showarticletitle{Mastering the game of Go without human knowledge}.
\newblock \bibinfo{journal}{\emph{Nature}} \bibinfo{volume}{550},
  \bibinfo{number}{7676} (\bibinfo{year}{2017}), \bibinfo{pages}{354}.
\newblock


\bibitem[\protect\citeauthoryear{Simon}{Simon}{1997}]%
        {simon1997models}
\bibfield{author}{\bibinfo{person}{Herbert~Alexander Simon}.}
  \bibinfo{year}{1997}\natexlab{}.
\newblock \bibinfo{booktitle}{\emph{Models of bounded rationality: Empirically
  grounded economic reason}}. Vol.~\bibinfo{volume}{3}.
\newblock \bibinfo{publisher}{MIT press}.
\newblock


\bibitem[\protect\citeauthoryear{Simonyan and Zisserman}{Simonyan and
  Zisserman}{2014}]%
        {simonyan2014very}
\bibfield{author}{\bibinfo{person}{Karen Simonyan} {and}
  \bibinfo{person}{Andrew Zisserman}.} \bibinfo{year}{2014}\natexlab{}.
\newblock \showarticletitle{Very deep convolutional networks for large-scale
  image recognition}.
\newblock \bibinfo{journal}{\emph{arXiv preprint arXiv:1409.1556}}
  (\bibinfo{year}{2014}).
\newblock


\bibitem[\protect\citeauthoryear{Stowell, Lyson, Saksono, Wurth, Jimison,
  Pavel, and Parker}{Stowell et~al\mbox{.}}{2018}]%
        {stowell2018designing}
\bibfield{author}{\bibinfo{person}{Elizabeth Stowell},
  \bibinfo{person}{Mercedes~C Lyson}, \bibinfo{person}{Herman Saksono},
  \bibinfo{person}{Rene{\'e}~C Wurth}, \bibinfo{person}{Holly Jimison},
  \bibinfo{person}{Misha Pavel}, {and} \bibinfo{person}{Andrea~G Parker}.}
  \bibinfo{year}{2018}\natexlab{}.
\newblock \showarticletitle{Designing and Evaluating mHealth Interventions for
  Vulnerable Populations: A Systematic Review}. In
  \bibinfo{booktitle}{\emph{Proceedings of the 2018 CHI Conference on Human
  Factors in Computing Systems}}. ACM, \bibinfo{pages}{15}.
\newblock


\bibitem[\protect\citeauthoryear{Szegedy, Ioffe, Vanhoucke, and Alemi}{Szegedy
  et~al\mbox{.}}{2017}]%
        {szegedy2017inception}
\bibfield{author}{\bibinfo{person}{Christian Szegedy}, \bibinfo{person}{Sergey
  Ioffe}, \bibinfo{person}{Vincent Vanhoucke}, {and}
  \bibinfo{person}{Alexander~A Alemi}.} \bibinfo{year}{2017}\natexlab{}.
\newblock \showarticletitle{Inception-v4, inception-resnet and the impact of
  residual connections on learning.}. In \bibinfo{booktitle}{\emph{AAAI}},
  Vol.~\bibinfo{volume}{4}. \bibinfo{pages}{12}.
\newblock


\bibitem[\protect\citeauthoryear{Szegedy, Liu, Jia, Sermanet, Reed, Anguelov,
  Erhan, Vanhoucke, and Rabinovich}{Szegedy et~al\mbox{.}}{2015}]%
        {szegedy2015going}
\bibfield{author}{\bibinfo{person}{Christian Szegedy}, \bibinfo{person}{Wei
  Liu}, \bibinfo{person}{Yangqing Jia}, \bibinfo{person}{Pierre Sermanet},
  \bibinfo{person}{Scott Reed}, \bibinfo{person}{Dragomir Anguelov},
  \bibinfo{person}{Dumitru Erhan}, \bibinfo{person}{Vincent Vanhoucke}, {and}
  \bibinfo{person}{Andrew Rabinovich}.} \bibinfo{year}{2015}\natexlab{}.
\newblock \showarticletitle{Going deeper with convolutions}. In
  \bibinfo{booktitle}{\emph{Proceedings of the IEEE conference on computer
  vision and pattern recognition}}. \bibinfo{pages}{1--9}.
\newblock


\bibitem[\protect\citeauthoryear{Szegedy, Vanhoucke, Ioffe, Shlens, and
  Wojna}{Szegedy et~al\mbox{.}}{2016}]%
        {szegedy2016rethinking}
\bibfield{author}{\bibinfo{person}{Christian Szegedy}, \bibinfo{person}{Vincent
  Vanhoucke}, \bibinfo{person}{Sergey Ioffe}, \bibinfo{person}{Jon Shlens},
  {and} \bibinfo{person}{Zbigniew Wojna}.} \bibinfo{year}{2016}\natexlab{}.
\newblock \showarticletitle{Rethinking the inception architecture for computer
  vision}. In \bibinfo{booktitle}{\emph{Proceedings of the IEEE conference on
  computer vision and pattern recognition}}. \bibinfo{pages}{2818--2826}.
\newblock


\bibitem[\protect\citeauthoryear{Tintarev and Masthoff}{Tintarev and
  Masthoff}{2007}]%
        {tintarev2007survey}
\bibfield{author}{\bibinfo{person}{Nava Tintarev} {and} \bibinfo{person}{Judith
  Masthoff}.} \bibinfo{year}{2007}\natexlab{}.
\newblock \showarticletitle{A survey of explanations in recommender systems}.
  In \bibinfo{booktitle}{\emph{Data Engineering Workshop, 2007 IEEE 23rd
  International Conference on}}. IEEE, \bibinfo{pages}{801--810}.
\newblock


\bibitem[\protect\citeauthoryear{Vtyurina and Fourney}{Vtyurina and
  Fourney}{2018}]%
        {vtyurina2018exploring}
\bibfield{author}{\bibinfo{person}{Alexandra Vtyurina} {and}
  \bibinfo{person}{Adam Fourney}.} \bibinfo{year}{2018}\natexlab{}.
\newblock \showarticletitle{Exploring the role of conversational cues in guided
  task support with virtual assistants}. In
  \bibinfo{booktitle}{\emph{Proceedings of the 2018 CHI Conference on Human
  Factors in Computing Systems}}. ACM, \bibinfo{pages}{208}.
\newblock


\bibitem[\protect\citeauthoryear{Wang, Oh, Wang, and Wiens}{Wang
  et~al\mbox{.}}{2018a}]%
        {Wang:2018:LCM:3219819.3220070}
\bibfield{author}{\bibinfo{person}{Jiaxuan Wang}, \bibinfo{person}{Jeeheh Oh},
  \bibinfo{person}{Haozhu Wang}, {and} \bibinfo{person}{Jenna Wiens}.}
  \bibinfo{year}{2018}\natexlab{a}.
\newblock \showarticletitle{Learning Credible Models}. In
  \bibinfo{booktitle}{\emph{Proceedings of the 24th ACM SIGKDD International
  Conference on Knowledge Discovery \&\#38; Data Mining}}
  \emph{(\bibinfo{series}{KDD '18})}. \bibinfo{publisher}{ACM},
  \bibinfo{address}{New York, NY, USA}, \bibinfo{pages}{2417--2426}.
\newblock
\showISBNx{978-1-4503-5552-0}
\urldef\tempurl%
\url{https://doi.org/10.1145/3219819.3220070}
\showDOI{\tempurl}


\bibitem[\protect\citeauthoryear{Wang, Oh, Wang, and Wiens}{Wang
  et~al\mbox{.}}{2018b}]%
        {wang2018learning}
\bibfield{author}{\bibinfo{person}{Jiaxuan Wang}, \bibinfo{person}{Jeeheh Oh},
  \bibinfo{person}{Haozhu Wang}, {and} \bibinfo{person}{Jenna Wiens}.}
  \bibinfo{year}{2018}\natexlab{b}.
\newblock \showarticletitle{Learning credible models}. In
  \bibinfo{booktitle}{\emph{Proceedings of the 24th ACM SIGKDD International
  Conference on Knowledge Discovery \& Data Mining}}. ACM,
  \bibinfo{pages}{2417--2426}.
\newblock


\bibitem[\protect\citeauthoryear{Wang and Benbasat}{Wang and Benbasat}{2007}]%
        {wang2007recommendation}
\bibfield{author}{\bibinfo{person}{Weiquan Wang} {and} \bibinfo{person}{Izak
  Benbasat}.} \bibinfo{year}{2007}\natexlab{}.
\newblock \showarticletitle{Recommendation agents for electronic commerce:
  Effects of explanation facilities on trusting beliefs}.
\newblock \bibinfo{journal}{\emph{Journal of Management Information Systems}}
  \bibinfo{volume}{23}, \bibinfo{number}{4} (\bibinfo{year}{2007}),
  \bibinfo{pages}{217--246}.
\newblock


\bibitem[\protect\citeauthoryear{Webber, Moffat, and Zobel}{Webber
  et~al\mbox{.}}{2010}]%
        {Webber:2010:RBO}
\bibfield{author}{\bibinfo{person}{William Webber}, \bibinfo{person}{Alistair
  Moffat}, {and} \bibinfo{person}{Justin Zobel}.}
  \bibinfo{year}{2010}\natexlab{}.
\newblock \showarticletitle{A Similarity Measure for Indefinite Rankings}.
\newblock \bibinfo{journal}{\emph{ACM Trans. Inf. Syst.}} \bibinfo{volume}{28},
  \bibinfo{number}{4}, Article \bibinfo{articleno}{20} (\bibinfo{date}{Nov.}
  \bibinfo{year}{2010}), \bibinfo{numpages}{38}~pages.
\newblock
\showISSN{1046-8188}
\urldef\tempurl%
\url{https://doi.org/10.1145/1852102.1852106}
\showDOI{\tempurl}


\bibitem[\protect\citeauthoryear{Xu, Ba, Kiros, Cho, Courville, Salakhudinov,
  Zemel, and Bengio}{Xu et~al\mbox{.}}{2015}]%
        {captioningxu2015showattention}
\bibfield{author}{\bibinfo{person}{Kelvin Xu}, \bibinfo{person}{Jimmy Ba},
  \bibinfo{person}{Ryan Kiros}, \bibinfo{person}{Kyunghyun Cho},
  \bibinfo{person}{Aaron Courville}, \bibinfo{person}{Ruslan Salakhudinov},
  \bibinfo{person}{Rich Zemel}, {and} \bibinfo{person}{Yoshua Bengio}.}
  \bibinfo{year}{2015}\natexlab{}.
\newblock \showarticletitle{Show, attend and tell: Neural image caption
  generation with visual attention}. In \bibinfo{booktitle}{\emph{International
  Conference on Machine Learning}}. \bibinfo{pages}{2048--2057}.
\newblock


\bibitem[\protect\citeauthoryear{Yamins, Hong, Cadieu, Solomon, Seibert, and
  DiCarlo}{Yamins et~al\mbox{.}}{2014}]%
        {yamins2014performance}
\bibfield{author}{\bibinfo{person}{Daniel~LK Yamins}, \bibinfo{person}{Ha
  Hong}, \bibinfo{person}{Charles~F Cadieu}, \bibinfo{person}{Ethan~A Solomon},
  \bibinfo{person}{Darren Seibert}, {and} \bibinfo{person}{James~J DiCarlo}.}
  \bibinfo{year}{2014}\natexlab{}.
\newblock \showarticletitle{Performance-optimized hierarchical models predict
  neural responses in higher visual cortex}.
\newblock \bibinfo{journal}{\emph{Proceedings of the National Academy of
  Sciences}} \bibinfo{volume}{111}, \bibinfo{number}{23}
  (\bibinfo{year}{2014}), \bibinfo{pages}{8619--8624}.
\newblock


\bibitem[\protect\citeauthoryear{Zarsky}{Zarsky}{2016}]%
        {zarsky2016trouble}
\bibfield{author}{\bibinfo{person}{Tal Zarsky}.}
  \bibinfo{year}{2016}\natexlab{}.
\newblock \showarticletitle{The trouble with algorithmic decisions: An analytic
  road map to examine efficiency and fairness in automated and opaque decision
  making}.
\newblock \bibinfo{journal}{\emph{Science, Technology, \& Human Values}}
  \bibinfo{volume}{41}, \bibinfo{number}{1} (\bibinfo{year}{2016}),
  \bibinfo{pages}{118--132}.
\newblock


\bibitem[\protect\citeauthoryear{Zheng, Liu, Ren, Ma, Chen, Yu, Xue, Chen, and
  Wang}{Zheng et~al\mbox{.}}{2017}]%
        {zheng2017hybrid}
\bibfield{author}{\bibinfo{person}{Nan-ning Zheng}, \bibinfo{person}{Zi-yi
  Liu}, \bibinfo{person}{Peng-ju Ren}, \bibinfo{person}{Yong-qiang Ma},
  \bibinfo{person}{Shi-tao Chen}, \bibinfo{person}{Si-yu Yu},
  \bibinfo{person}{Jian-ru Xue}, \bibinfo{person}{Ba-dong Chen}, {and}
  \bibinfo{person}{Fei-yue Wang}.} \bibinfo{year}{2017}\natexlab{}.
\newblock \showarticletitle{Hybrid-augmented intelligence: collaboration and
  cognition}.
\newblock \bibinfo{journal}{\emph{Frontiers of Information Technology \&
  Electronic Engineering}} \bibinfo{volume}{18}, \bibinfo{number}{2}
  (\bibinfo{year}{2017}), \bibinfo{pages}{153--179}.
\newblock


\end{thebibliography}

\end{document}